\documentclass{article}

\usepackage{PRIMEarxiv}

\usepackage[utf8]{inputenc} 
\usepackage[T1]{fontenc}    
\usepackage{hyperref}       
\usepackage{url}            
\usepackage{booktabs}       
\usepackage{amsfonts}       
\usepackage{nicefrac}       
\usepackage{microtype}      
\usepackage{lipsum}
\usepackage{multirow}
\usepackage{fancyhdr}       
\usepackage{graphicx}     
\usepackage{comment}  
\usepackage{longtable}   
\usepackage{caption}
\usepackage{tabularx}     
\usepackage{pdflscape}
\usepackage{rotating}
\usepackage{algorithm}
\usepackage{algorithmic}
\usepackage{amsmath}
\usepackage{float}
\usepackage{subcaption}
\usepackage{orcidlink}
\graphicspath{{media/}}     

\title{Benchmarking Scientific Machine Learning Models for Air Quality Data
\thanks{\textit{Corresponding Author Email}: 
\textbf{sahara.ali@unt.edu}} 
}

\author{
	Khawja Imran Masud\orcidlink{0009-0005-6656-5914} \\
	University of North Texas\\
	76205 Texas, USA \\
	\texttt{KhawjaImran.Masud@unt.edu} \\
	\And
	Venkata Sai Rahul Unnam \\
	University of North Texas\\
	76205 Texas, USA \\
	\texttt{VenkataSaiRahulUnnam@my.unt.edu} \\
	\And
	Sahara Ali\orcidlink{0000-0002-8578-948X} \\
	University of North Texas\\
	76205 Texas, USA \\
	\texttt{sahara.ali@unt.edu} \\
}

\begin{document}
\maketitle

\begin{abstract}
Accurate air quality index (AQI) forecasting is essential for the protecting public health in rapidly growing urban regions, and the practical model evaluation and selection are often challenged by the lack of rigorous, region-specific benchmarking on standardized datasets. Physics-guided machine learning and deep learning models could be a good and effective solution to resolve such issues with more accurate and efficient AQI forecasting. This research study presents an explainable and comprehensive benchmark that enables a guideline and proposed physics-guided best model by benchmarking classical time-series, machine-learning, and deep-learning approaches for multi-horizon AQI forecasting in North Texas (Dallas County). Using publicly available U.S. Environmental Protection Agency (EPA) daily observations of air quality data from 2022 to 2024, we curate city-level time series for PM$_{2.5}$ and O$_3$ by aggregating station measurements and constructing lag-wise forecasting datasets for $LAG \in \{1,7,14,30\}$ days. For benchmarking the best model, linear regression (LR), SARIMAX, multilayer perceptrons (MLP), and LSTM networks are evaluated with the proposed physics-guided variants (MLP+Physics and LSTM+Physics) that incorporate the EPA breakpoint-based AQI formulation as a consistency constraint through a weighted loss. Experiments using chronological train-test splits and error metrics MAE, RMSE showed that deep-learning models outperform simpler baselines, while physics guidance improves stability and yields physically consistent pollutant with AQI relationships, with the largest benefits observed for short-horizon prediction and for PM$_{2.5}$ and O$_3$. Overall, the results provide a practical reference for selecting AQI forecasting models in North Texas and clarify when lightweight physics constraints meaningfully improve predictive performance across pollutants and forecast horizons. 
\end{abstract}


\section{Introduction}

Air pollution is one of the most pressing global challenges today; in particular, elevated concentrations of fine particulate matter (PM$_{2.5}$) and ozone (O$_3$) pose substantial risks to human health and environmental well-being \cite{zhang2024association, alexeeff2023association, sangkham2024update}. These global concerns are especially evident in the North Texas region, which has been experiencing significant population growth, economic expansion, and massive air quality issues due to the industrial revolution, urbanization, and increased vehicle traffic \cite{fortworthreport2025worst, siegmund2025development}. AQI forecasting is a significant metric to evaluate the pollution levels that helps communicate related health risks to the public and policymakers \cite{epa2024_aqi_tech_doc, horn2024air}. Over decades, approaches to AQI and pollutant forecasting have evolved from simple statistical models to advanced machine learning (ML), deep learning (DL), and, more recently, scientific ML.

In early work, researchers leveraged primarily statistical regressions and the time-series models to forecast pollution levels \cite{goyal2006statistical, taheri2016statistical, zhang2017forecasting, liu2018time}, but their finding capability was much limited; such a shortcoming then motivated the transition towards more flexible, data-driven approaches. With the growth of multi-pollutant monitoring networks and improved data availability, machine learning (ML) emerged as a compelling alternative to classical statistical methods. ML techniques, including Random Forest, Gradient Boosting Machines, Support Vector Regression, and CatBoost, demonstrated substantial performance gains by modeling nonlinear relationships between pollutants, meteorology, and emission activities \cite{li2022random, ravindiran2023air, natarajan2024optimized, yenkikar2025explainable}. These approaches enhanced the predicted stability and interpretability compared to standalone statistical models; however, to improve the accuracy in AQI prediction, researchers have developed more advanced deep-learning models, which rely solely on historical data correlations and find it difficult to grasp deeper temporal connections \cite{sarkar2022air, elbaz2023real, nguyen2024predicting, hu2025geographically, utku2025advancing}. These models are primarily data-driven and may violate physical consistency under unknown circumstances, despite their good predictive performance. To address these limitations,  physics-informed neural networks (PINNs) \cite{raissi2019physics} were introduced as a way to enforce physical consistency within machine-learning and deep-learning models and lightly applied in air quality and environmental data \cite{shi2025phy, li2025knowledge, chandrashekar2025hybrid}.

However, no prior study has systematically benchmarked the effect of integrating simple AQI formulation physics across multiple model families, pollutants (PM$_{2.5}$) and ozone (O$_3$), and temporal lag structures. Researchers and policymakers often find it difficult to select the most appropriate model for a specific objective, such as ozone forecasting, particulate matter (PM$_{2.5}$) prediction, or data imputation, due to the absence of direct comparative studies conducted on a standardized dataset. This limitation highly motivated us to evaluate how incorporating EPA-style piecewise AQI mapping into diverse ML and DL architectures influences predictive performance, robustness, and physical consistency across lag-wise (1, 7, 14, and 30 days) forecasting scenarios. This research addresses this gap by benchmarking state-of-the-art scientific machine learning methods on a North Texas air quality dataset. The main contributions are threefold:

\begin{enumerate}

\item The curation of a comprehensive, multi-source environmental dataset for the Dallas region; 

\item Proposed and implemented physics-informed SciML models MLP+Physics and LSTM+Physics for air quality time-series forecasting; and 

\item A comparative evaluation of these methods using multiple performance metrics.

\end{enumerate}

The Related Study section presents an organized summary of existing literature in the area pertinent to the current research in an attempt to place the proposed research in the context of the broader air quality modeling and forecasting literature. It first reviews existing approaches to air quality monitoring and prediction, including traditional statistical models, machine learning techniques, and recent Scientific Machine Learning (SciML) frameworks. The section then draws attention to commonly used datasets and sources of data, motivating the choice of design described in the Dataset Description section. Next, it discusses methodological trends and limitations in existing studies that help inform model selection and experimental design, which are presented in the Methodology section. Finally, gaps in comparative evaluation and region-specific benchmarking identified in the literature establish the need for the Results and Comparative Analysis section and motivate the broader discussion and conclusions drawn in the final sections of this paper.

\section{Related Study}
Air pollution, in particular, fine particulates (PM$_{2.5}$) and ozone (O$_3$) at high concentrations, is a major threat to human health and the environment \cite{zhang2024association, alexeeff2023association, sangkham2024update}. Accurately predicting the Air Quality Index (AQI) or concentrations of important pollutants is important to public health warnings, urban planning, and environmental policy. Over decades, different approaches have been used to forecast AQI and pollutants, from basic statistical models to advanced ML, DL, and, more recently, physics-informed scientific ML.

In early work, researchers relied primarily on statistical regressions and time-series models to forecast pollution levels or AQI \cite{goyal2006statistical, taheri2016statistical, zhang2017forecasting, liu2018time}. These models generally used to use past concentrations of pollutants and meteorological variables to predict future air quality. While effective in capturing short-term linear relationships, they struggled with non-linear dynamics, sudden spikes, and complex pollutant–meteorology interactions typical in urban environments. However, their predictive capability was limited, a shortcoming that motivated the transition towards more flexible, data-driven approaches.

With the development of multi-pollutant monitoring networks and improved data availability, machine learning (ML) \cite{dadhich2021machine}  emerged as a compelling alternative to classical statistical methods. ML techniques, including Random Forest,
Gradient Boosting Machines, Support Vector Regression, and CatBoost, demonstrated substantial performance gains by modeling nonlinear relationships between pollutants, meteorology, and emission activities. Li et al. \cite{li2022random} have used a Random Forest model to estimate PM$_{2.5}$ concentration on a daily basis in the Huaihai Economic Zone based on satellite AOD, weather data, and spatial-temporal features. The model performed well ($R^2 = 0.85$, RMSE = 14.63 µg/m$^3$) and generated 1-km maps of pollution and trends over time. However, because Random Forest cannot capture complex nonlinear or
long-range temporal patterns, its ability to model fast-changing atmospheric conditions is limited, showing the need for more advanced deep-learning methods.  Another research study by Ravindiran et al. \cite{ravindiran2023air} used several machine-learning models predict AQI in Visakhapatnam. CatBoost gave the best results, with very high accuracy ($R^2 \approx 0.9998$, RMSE $\approx 0.76$), and performed much better than traditional methods. This indicates that ML models have the ability to capture the complex relationships in AQI data. 

However, these models remain statistical and do not involve the actual physics of the atmosphere.Because of this, they may work well on past data but can be less reliable during unusual weather, climate changes, or when data is limited. Similarly, Natarajan et al. \cite{natarajan2024optimized} developed an optimized ML framework that integrates Grey Wolf Optimization with a Decision Tree regressor and achieved high accuracies across major Indian cities (88.98\%–97.68\%), outperforming SVR and Random Forest through improved feature selection and class-imbalance handling. In addition, Yenkikar et al. \cite{yenkikar2025explainable} developed a hybrid Random Forest–ARIMA model for AQI forecasting, where Random Forest captured nonlinear pollutant interactions while ARIMA modeled residual temporal patterns. This hybrid approach improved predictive stability and interpretability compared to standalone ML models, although it remained limited by its reliance on historical correlations. Yet, despite these gains, this model also relies entirely on historical data correlations and struggles to capture deeper temporal dependencies, suggesting that more advanced deep-learning models may be necessary to enhance robustness and generalization in AQI prediction.

Moreover, the recent deep-learning literature shows that more expressive temporal and spatio-temporal models can substantially improve air-quality forecasting beyond classical ML. Early hybrid work such as Sarkar et al. \cite{sarkar2022air}, who combined LSTM and GRU for AQI prediction in Delhi and reported lower MAE and higher $R^2$ than stand-alone DL and ML baselines, already indicated that recurrent architectures can better capture non-linear pollutant dynamics over time. Building on this, Elbaz et al. \cite{elbaz2023real} proposed a 3D-CNN-GRU model with an attention mechanism in Chemosphere to generate multi-horizon forecasts from image-based and environmental inputs, showing that jointly modeling spatial texture and temporal evolution yields more accurate PM$_{2.5}$ predictions than conventional CNN or RNN models. More recently, several high-quality studies from 2024–2025 have explored increasingly sophisticated hybrid and attention-based architectures. For instance, Nguyen et al. \cite{nguyen2024predicting} developed an Attention-CNN combined with a QPSO-optimized LSTM framework that significantly reduced RMSE and MAE for daily AQI prediction in Seoul, demonstrating the value of attention and meta-heuristic tuning in capturing fast-changing atmospheric patterns. Similarly, Hu et al. \cite{hu2025geographically} introduced a geographically aware CNN–LSTM–KAN hybrid model that integrates climatic and topographic differentiation, and demonstrated a 23.6–59.6\% reduction in RMSE and $R^2$ of 0.92–0.99 across five geographically and climatically distinct Chinese cities, outperforming standard LSTM and CNN–LSTM baselines. In addition, Utku et al. \cite{utku2025advancing} introduced a lightweight CNN–RNN hybrid evaluated across India, Milan, and Frankfurt, where it outperformed several ML and DL baselines. Although the study did not explicitly analyze emission-regime shifts, the model’s stable performance across diverse locations suggests potential for broader generalization. Collectively, these deep-learning advances highlight substantial gains in modeling non-stationary pollutant behavior and complex spatio-temporal dependencies. However, despite their predictive power, these architectures remain purely data-driven and do not incorporate atmospheric physics, pollutant transformation rules, or AQI formulation principles. As a result, their outputs may violate physical consistency under unseen conditions or regime shifts emphasizing the need for physics-informed or scientific machine-learning approaches that blend data with domain knowledge.

To mitigate these limitations, physics-based solutions were proposed in the form of physics-informed neural networks (PINNs) to impose physical consistency on machine learning and deep learning models. Instead of using only historical correlation, PINNs incorporate governing equations and domain constraints directly into the learning process \cite{raissi2019physics}.  Building on this foundation, recent studies have begun applying physics-informed and scientific machine-learning approaches to air-quality problems. For instance, Shi et al. \cite{shi2025phy} proposed Phy-APMR, a model that couples a neural network with a PDE-based pollutant-transport formulation to reconstruct fine-grained concentration maps from sparse mobile sensors. Their framework achieved roughly 15\% lower reconstruction errors and up to 84\% faster training than state-of-the-art baselines, illustrating the value of embedding physical propagation constraints under data sparsity. Similarly, Li et al. \cite{li2025knowledge} developed a knowledge-informed deep-learning model for joint prediction of multiple pollutants, reducing bias and improving generalization; however, because the method does not incorporate explicit transport equations, its physical grounding remains limited under atypical atmospheric regimes. In parallel, Chandrashekar et al. \cite{chandrashekar2025hybrid} introduced AirSense-X, a hybrid PINN-inspired and XAI-based model that attained approximately 98\% accuracy in AQI classification. Yet the absence of explicit physical equations and the focus on categorical AQI prediction, rather than forecasting future concentrations, raise questions about how truly “physics-informed” the model is and whether it can generalize under evolving atmospheric conditions.

Overall, no prior study has systematically benchmarked the effect of integrating simple AQI formulation physics across multiple model families, pollutants (PM$_{2.5}$) and ozone (O$_3$), and temporal lag structures. This gap motivates the present work, which evaluates how incorporating EPA-style piecewise AQI mapping into diverse ML and DL architectures influences predictive performance, robustness, and physical consistency across lag-1, lag-7, lag-14, and lag-30 forecasting scenarios. Through this benchmarking analysis, the study seeks to clarify when and how physics-based constraints meaningfully enhance air-quality prediction, and whether such improvements are consistent across different pollutants and model complexities. Table~\ref{tab:literature_reduced} summarizes representative ML, DL, hybrid, and physics-informed approaches used in recent air-quality prediction studies, highlighting their architectures, performance, and key limitations.

\begin{table*}[htbp]
	\centering
	\footnotesize
	\renewcommand{\arraystretch}{1.3}
	\resizebox{\textwidth}{!}{%
		\begin{tabular}{p{2.4cm} p{3.0cm} p{2.8cm} p{2.3cm} p{3.0cm} p{3.8cm}}
			\toprule
			\textbf{Related Study} &
			\textbf{Model Architecture} &
			\textbf{Datasets} &
			\textbf{Task} &
			\textbf{Results} &
			\textbf{Tentative Research Gap} \\
			\midrule

			Li et al. \cite{li2022random} &
			Random Forest regression &
			Huaihai Economic Zone, China (2000--2020) &
			Daily PM$_{2.5}$ estimation &
			$R^2 = 0.85$, RMSE = 14.63~$\mu$g/m$^3$, MAE = 10.03~$\mu$g/m$^3$ &
			No physics; cannot capture long-range temporal interactions. \\
			\\
			Ravindiran et al. \cite{ravindiran2023air} &
			CatBoost / ensemble ML &
			Visakhapatnam, India (2017--2022) &
			AQI forecasting &
			$R^2 \approx 0.9998$, RMSE $\approx 0.76$ &
			Extremely high fit; may overfit; relies entirely on historical correlations. \\
			\\
			Yenkikar et al. \cite{yenkikar2025explainable} &
			Random Forest + ARIMA residual modeling &
			India (station-level data) &
			AQI forecasting &
			MSE $\approx$ 508.46; $R^2 \approx 0.94$ &
			No DL; limited temporal modeling; no physical knowledge. \\
			\\
			Sarkar et al. \cite{sarkar2022air} &
			LSTM--GRU hybrid &
			Delhi, India &
			AQI prediction &
			MAE = 36.11; $R^2 = 0.84$ &
			Single-city; no spatial modeling; fully data-driven. \\
			\\
			Elbaz et al.\cite{elbaz2023real} &
			3D-CNN + GRU + Attention &
			China (regional / national) &
			Multi-horizon PM$_{2.5}$ forecasting &
			Outperformed 2D-CNN, 3D-CNN, GRU baselines in RMSE / MAE &
			Needs large image + met datasets; computationally heavy; no explicit physics. \\
			\\
			Hu et al. \cite{hu2025geographically} &
			CNN--LSTM--KAN hybrid &
			Five Chinese cities &
			AQI / pollutant forecasting &
			RMSE reduced 23.6--59.6\%; $R^2 = 0.92$--$0.99$ &
			High complexity; generalization beyond studied cities untested. \\
			\\
			Shi et al. \cite{shi2025phy} &
			Phy-APMR (NN + pollutant-transport PDE + ASUS) &
			Multiple Chinese cities &
			Fine-grained pollution map reconstruction &
			$\sim$15\% lower error; $\sim$84\% faster convergence &
			Requires mobile sensors; focuses on mapping, not forecasting; PDE assumptions limit transferability. \\
			\\
			Chandrashekar et al. \cite{chandrashekar2025hybrid} &
			PINN + XAI (AQI classification) &
			India (multi-city) &
			AQI classification &
			Accuracy $\approx 98$\%; precision 97\%; recall 95\%; F1 = 0.96 &
			No explicit PDE physics; classification only, not forecasting. \\
			\bottomrule
		\end{tabular}%
	}
	\caption{Comparison of major ML, DL, hybrid, and physics-informed approaches for air-quality prediction and forecasting.}
	\label{tab:literature_reduced}
\end{table*}

\section{Datasets Description}
\label{sec:dataset_description}

\subsection{Data Collection}
In this research study, a publicly available air quality data set is collected from the United States Environmental Protection Agency (EPA) through the outdoor daily air quality data portal \cite{EPA_AirData_Daily}. The dataset includes three consecutive years of observations, 2022, 2023, and 2024, for benchmarking the physics impact on AQI prediction.  To analyze the air quality of the highly polluted zone in Texas, Dallas County, has been selected, and data has been collected from a total of three stations (Convention Center, Dallas Hinton, and Dallas Bexar Street). For each year, we downloaded the daily records for fine particulate matter Particulate Matters (PM$_{2.5}$) and Ozone (O$_3$), and the Air Quality Index (AQI) from all EPA monitoring stations operating within the Dallas county. 

\subsection{Data Aggregation}
Each raw yearly dataset contains daily pollutant measurements collected independently at multiple monitoring sites. To obtain a representative city-level estimate of daily air quality, all station observations for the same day were averaged. This produced a single time-series file per year containing the daily mean PM$_{2.5}$ concentration and the corresponding daily mean AQI value. The three yearly files were then merged into one continuous multi-year dataset covering the period January 2022 to December 2024. Similarly, we collected data for the ozone (O$_3$) and then merged it into one continuous multi-year dataset covering the period January 2022 to December 2024 as well. Before analysis, the merged datasets were cleaned by converting date formats to a unified structure, removing incomplete records, and ensuring consistent column naming. This resulted in a high-quality daily datasets for PM$_{2.5}$ and O$_3$ containing three primary variables: DATE, Daily\_Mean\_PM or Daily\_Mean\_Ozone, and Daily\_AQI\_Value. Table~\ref{tab:base-dataset-summary} summarizes the base PM\textsubscript{2.5} and O\textsubscript{3} datasets used to construct multi-horizon AQI targets. The datasets contain 1094 and 1091 daily observations, respectively, covering the period 2022–2024. Each record includes the daily mean pollutant concentration and the corresponding EPA AQI value. The datasets exhibit substantial variability in both pollutant levels and AQI, providing a robust foundation for deep learning–based forecasting experiments.

\begin{table}[htbp]
	\centering
	\caption{Summary statistics of the base PM$_{2.5}$ and O$_3$ datasets used for constructing multi-horizon AQI targets.}
	\vspace{2mm}
	\label{tab:base-dataset-summary}
	\begin{tabular}{lcc}
		\toprule
		\textbf{Statistic} 
		& \textbf{PM$_{2.5}$ Dataset (2022--2024)} 
		& \textbf{O$_3$ Dataset (2022--2024)} \\ 
		\toprule
		Number of daily records 
		& 1094 
		& 1091 \\
		
		AQI (min) 
		& 6.00 
		& 2.67 \\
		
		AQI (max) 
		& 124.00 
		& 168.33 \\
		
		AQI (mean $\pm$ std) 
		& 46.18 $\pm$ 14.90 
		& 41.82 $\pm$ 20.78 \\
		
		Pollutant value (min) 
		& 0.45 $\mu$g/m$^3$ 
		& 0.0027 ppm \\
		
		Pollutant value (max) 
		& 44.85 $\mu$g/m$^3$ 
		& 0.0927 ppm \\
		
		Pollutant (mean $\pm$ std) 
		& 9.62 $\pm$ 4.90 $\mu$g/m$^3$ 
		& 0.0415 $\pm$ 0.0138 ppm \\
		
		\bottomrule
	\end{tabular}
\end{table}

\subsection{Lag-Wise Dataset Transformation for Multi-Horizon Forecasting}

For benchmarking for AQI forecasting, the individual multi-year datasets were further converted into a lag-based prediction structure. In this formulation the target variable is the future value of the AQI, shifted up by a specified number of days. The forecasting horizon is set to $LAG = 1$, $LAG = 7$, $LAG = 14$, and $LAG = 30$, and the dataset matches the pollutant values measured on each day with the AQI value of the following lag days. This transformation introduces an additional variable, AQI\_Targeted\_Value\_LAG\_L, representing the AQI value $L$ days ahead of each observation. Rows lacking valid future AQI values (such as the final $L$ days of the dataset) were removed to preserve dataset completeness. Each LAG-specific dataset therefore contains well-aligned feature target pairs comprising with current day's pollutant concentration (PM$_{2.5}$ or O$_3$), Current day's daily AQI value, and the AQI value $L$ days in the future target. A total of four datasets have been created for PM$_{2.5}$ with LAG sizes of 1, 7, 14, and 30, and another four datasets have been created for O$_3$ with the same LAG values. Datasets preparation techniques is showed in the Algorithm \ref{alg:lag_aqi}. This LAG-wise technique enables rigorous forecasting experiments for various time horizons and supports the evaluation of physical influences on AQI predictability. The final processed datasets are clean, temporally consistent, and well suited for machine learning and physics-informed modeling, forming the foundation of the predictive analysis conducted in this study.

\begin{algorithm}[h]
	\caption{Construction of Multi-Year Lag-based AQI Prediction Dataset}
	\label{alg:lag_aqi}
	\begin{algorithmic}[1]
		
		\renewcommand{\algorithmicrequire}{\textbf{Input:}}
		\renewcommand{\algorithmicensure}{\textbf{Output:}}
		
		\REQUIRE Years $\mathcal{Y}=\{2022,2023,2024\}$; Collect EPA Daily Data $\{\mathrm{file}_y : y\in\mathcal{Y}\}$; forecast horizon $LAG \in \mathbb{N}$
		\ENSURE Lag-wise dataset $\mathcal{D}^{(LAG)}$ for Lag-day-ahead AQI forecasting
		\vspace{2mm}
		\STATE \textbf{Stage 1: Multi-Year Station-Averaged EPA Dataset}
		
		\STATE Initialize list $\mathcal{D}_{avg} \gets [~]$
		
		\FOR{each year $y \in \mathcal{Y}$}
		\STATE $D_y \gets \text{LoadCSV}(\text{file}_y)$
		\STATE Convert $D_y[\text{Date}]$ to datetime
		\STATE $D_y \gets \text{DropNA}(D_y,\{\text{Date}, \text{Conc}, \text{AQI}\})$
		\STATE $D_y \gets \text{SortByDate}(D_y)$
		
		\FOR{each unique date $t$ in $D_y$}
		\STATE $\bar{c}_y(t) \gets \text{Mean}\{ r.\text{Conc} : r.\text{Date}=t \}$
		\STATE $\bar{a}_y(t) \gets \text{Mean}\{ r.\text{AQI} : r.\text{Date}=t \}$
		\ENDFOR
		
		\STATE $D_y^{avg} \gets \{ (t, \bar{c}_y(t), \bar{a}_y(t))\}_{t}$
		\STATE Append $D_y^{avg}$ to list $\mathcal{D}_{avg}$
		\ENDFOR
		
		\STATE $D^{multi} \gets \text{Concat}(\mathcal{D}_{avg})$
		\vspace{2mm}
		\STATE \textbf{Stage 2: Lag-based Dataset Preparation}
		
		\STATE Load combined CSV into data frame $df$
		
		\STATE $df \gets \mathrm{ResetIndex}(df)$
		\STATE $data \gets df[\mathrm{DATE}, \mathrm{Daily\_Mean\_PM}, \mathrm{Daily\_AQI\_Value}]$
		
		\STATE Choose forecast horizon $LAG \ge 1$
		
		\FOR{$i = 1$ to number of rows in $data$}
		\IF{$i + LAG \le$ number of rows}
		\STATE $\mathrm{AQI\_Targeted\_Value\_LAG\_L}[i] \gets \mathrm{Daily\_AQI\_Value}[i + LAG]$
		\ELSE
		\STATE $\mathrm{AQI\_Targeted\_Value\_LAG\_L}[i] \gets \mathrm{NaN}$
		\ENDIF
		\ENDFOR
		
		\STATE Append column $\mathrm{AQI\_Targeted\_Value\_LAG\_L}$ to $data$
		
		\STATE $\mathcal{D}^{(LAG)} \gets data[\mathrm{DATE}, \mathrm{Daily\_Mean\_PM}, \mathrm{Daily\_AQI\_Value}, \mathrm{AQI\_Targeted\_Value\_LAG\_L}]$
		\RETURN $\mathcal{D}^{(LAG)}$
		
	\end{algorithmic}
\end{algorithm}

To illustrate the structure of the lag-wise datasets used for forecasting, Tables~\ref{tab:merged_side_by_side_PM_O3} present sample records for PM$_{2.5}$ and O$_3$, respectively. It lists the current-day features, daily mean pollutant concentration and corresponding AQI value alongside four future AQI targets generated using different forecasting horizons (LAG = 1, 7, 14, and 30 days). These examples demonstrate how each observation is aligned with multiple future AQI values, enabling the development and evaluation of multi-horizon air-quality prediction models for both PM$_{2.5}$ and O$_3$. 

\begin{table*}[htbp]
	\centering
	\caption{First 10 samples of the lag-wise PM$_{2.5}$ and O$_3$ datasets across multiple forecasting horizons.}
	
	\begin{tabular}{l |cc|cccc | cc|cccc}
		\toprule
		\multicolumn{7}{c|}{\textbf{PM$_{2.5}$ Data}} &
		\multicolumn{6}{c}{\textbf{O$_3$ Data}} \\
		\cmidrule(lr){1-7} \cmidrule(lr){8-13}
		\textbf{DATE} & \textbf{PM$_{2.5}$} & \textbf{AQI} &
		\textbf{L1} & \textbf{L7} & \textbf{L14} & \textbf{L30} &
		\textbf{O$_3$} & \textbf{AQI} &
		\textbf{L1} & \textbf{L7} & \textbf{L14} & \textbf{L30} \\
		\midrule
		01-01-22 & 6.10  & 34.00 & 26.00 & 61.00 & 27.00 & 19.00 &
		0.0250 & 23.00 & 30.33 & 15.67 & 26.67 & 31.33 \\
		02-01-22 & 4.60  & 26.00 & 52.00 & 23.00 & 55.00 & 24.50 &
		0.0323 & 30.33 & 27.67 & 24.33 & 29.00 & 33.00 \\
		03-01-22 & 9.60  & 52.00 & 29.00 & 31.00 & 64.25 & 21.00 &
		0.0297 & 27.67 & 33.33 & 22.00 & 28.67 & 27.67 \\
		04-01-22 & 5.30  & 29.00 & 32.67 & 49.00 & 37.00 & 6.00  &
		0.0360 & 33.33 & 30.67 & 26.67 & 40.67 & 28.67 \\
		05-01-22 & 5.90  & 32.67 & 39.00 & 35.00 & 44.00 & 41.33 &
		0.0330 & 30.67 & 23.00 & 29.67 & 23.33 & 34.00 \\
		06-01-22 & 7.10  & 39.00 & 42.00 & 43.00 & 9.00  & 64.00 &
		0.0250 & 23.00 & 23.33 & 28.67 & 26.00 & 34.00 \\
		07-01-22 & 7.50  & 42.00 & 61.00 & 52.67 & 17.00 & 32.00 &
		0.0253 & 23.33 & 15.67 & 41.33 & 26.50 & 36.33 \\
		08-01-22 & 14.47 & 61.00 & 23.00 & 27.00 & 55.00 & 42.33 &
		0.0167 & 15.67 & 24.33 & 26.67 & 25.00 & 30.33 \\
		09-01-22 & 4.20  & 23.00 & 31.00 & 55.00 & 50.33 & 25.50 &
		0.0263 & 24.33 & 22.00 & 29.00 & 33.50 & 34.33 \\
		10-01-22 & 5.60  & 31.00 & 49.00 & 64.25 & 26.00 & 50.00 &
		0.0237 & 22.00 & 26.67 & 28.67 & 18.00 & 32.33 \\
		\bottomrule
	\end{tabular}
	\label{tab:merged_side_by_side_PM_O3}
\end{table*}

\subsection{Data Pre-processing}
To fit the data into the different targeted model, rows containing missing features or missing future-day targets were validated to ensure complete training examples. After constructing these final lag-wise datasets, all samples were split chronologically using an 80–20 ratio, where the first portion formed the training set and the remaining future portion formed the test set. This prevented information from future days from leaking into the training stage and ensured that all models were evaluated under realistic forecasting conditions.

As, different model families required slightly different preparation steps. Linear regression used the raw, unscaled pollutant and AQI values. SARIMAX used the AQI series as the main time-series input and included a one-day-lagged pollutant value as an exogenous regressor; the series was set to daily frequency, forward-filled when necessary, and split without scaling. In contrast, MLP and MLP+Physics models applied standardization to input features using statistics computed only from the training set, while keeping AQI targets in their real units. LSTM and LSTM+Physics models used min-max normalization for both inputs and targets to support stable gradient-based training; after prediction, AQI outputs were inverse-transformed back to real physical units. This unified and carefully controlled preprocessing pipeline ensured consistency across model families, avoided data leakage, and preserved the interpretability of AQI predictions.

\section{Methodology}
\label{sec:methodology}

\begin{figure*}[htbp]
	\centering
	\includegraphics[scale=0.55]{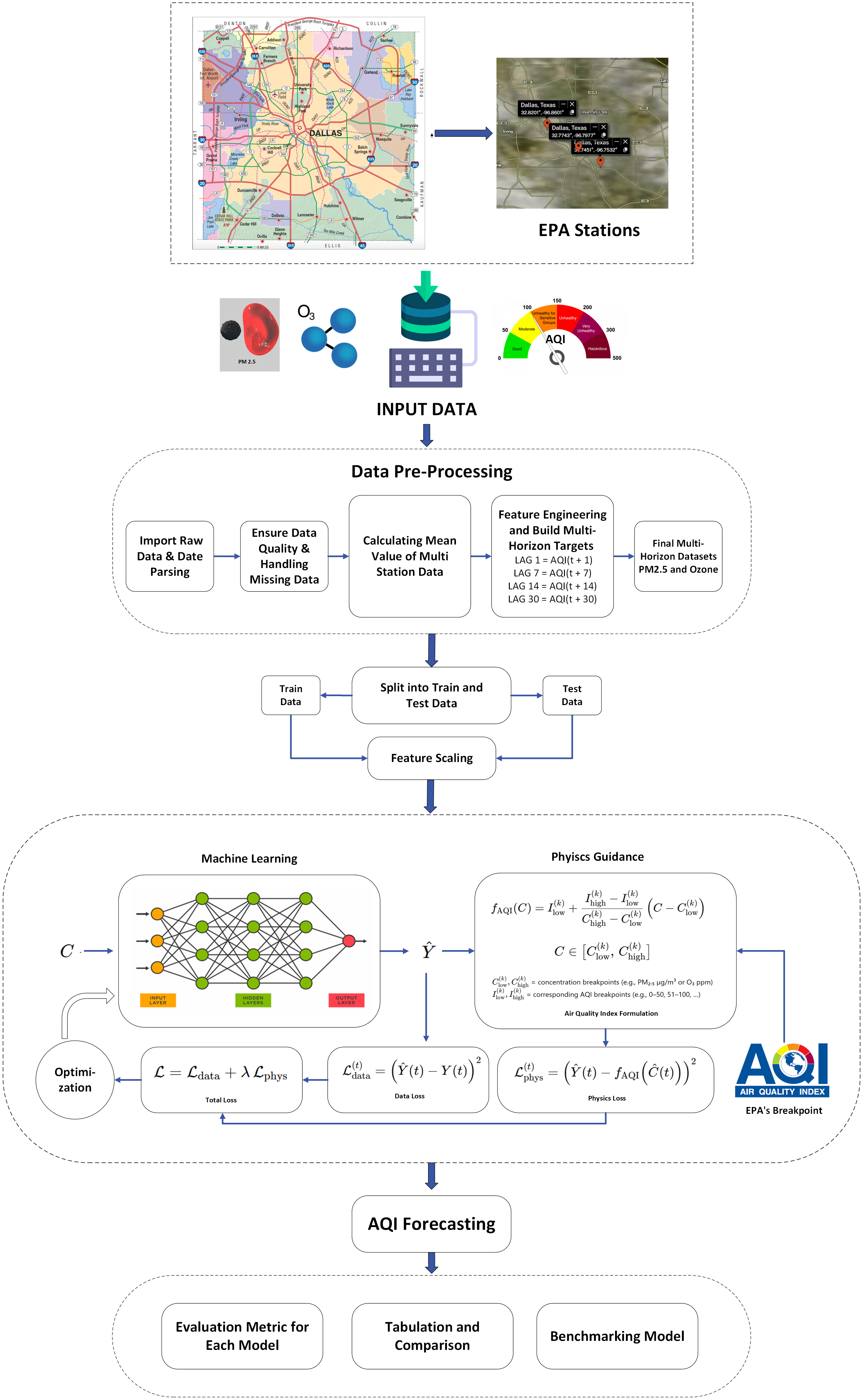}
	\caption{
		Proposed conceptual model for Physics Guided Scientific Machine Learning for Air Quality Index Forecasting.
	}
	\label{fig:ProposedModelSciMLAQI}
\end{figure*}

A structured scientific machine learning workflow has been designed and followed to integrate the environment data processing, create supervised multi-horizon datasets for the forecast, develop models for benchmarking, integrate physics into the model learning, and perform a detailed comparative evaluation for benchmarking the air quality index for Dallas County. The overall conceptual model is that starting from raw pollutant and AQI measurements from EPA monitoring stations, the data is passed through a series of data preparation, feature engineering, model training, and performance benchmarking steps, which are reflected in the workflow diagram in Figure \ref{fig:ProposedModelSciMLAQI}. 

\begin{algorithm}[htbp]
	\caption{Generalized Multi-Model Lag-Wise AQI Forecasting with and without Physics Guidance}
	\label{alg:multi_model_benchmark}
	\begin{algorithmic}[1]
		
		\renewcommand{\algorithmicrequire}{\textbf{Input:}}
		\renewcommand{\algorithmicensure}{\textbf{Output:}}
		
		\REQUIRE
		Lag-wise datasets $\mathcal{D}^{(p,LAG)}$ constructed using Algorithm~\ref{alg:lag_aqi} for pollutants $p \in \{\mathrm{PM}_{2.5}, \mathrm{O}_3\}$ and horizons $LAG \in \{1,7,14,30\}$;\\
		Model set $\mathcal{M} = \{\mathrm{LR}, \mathrm{SARIMA}, \mathrm{MLP}, \mathrm{MLP+Phys}, \mathrm{LSTM}, \mathrm{LSTM+Phys}\}$;\\
		Train--test split ratio $\alpha$ (e.g., 0.8); EPA AQI breakpoint tables $\mathcal{B}$; loss weights $\lambda_{\mathrm{data}}$, $\lambda_{\mathrm{phys}}$.
		
		\ENSURE
		Trained models and evaluation metrics (MAE, RMSE, $R^2$) for each $(p, LAG, m)$.
		
		\vspace{1mm}
		\STATE \textbf{Stage 1: Dataset Pre-Processing}
		\STATE Use Algorithm~\ref{alg:lag_aqi} to construct the LAG-wise dataset $\mathcal{D}^{(p,LAG)}$.
		\STATE Split each $\mathcal{D}^{(p,LAG)}$ chronologically into training (first $\alpha$) and testing (remaining $1-\alpha$).
		\STATE Fit a scaler on training features/targets only (for ML/DL models) and transform both sets. SARIMA uses the raw AQI series without scaling.
		
		\vspace{1mm}
		\STATE \textbf{Stage 2: Model Definition}
		\FOR{each model type $m \in \mathcal{M}$}
		\IF{$m =$ LR}
		\STATE Define linear regression model $y = \beta_0 + \beta_1\text{Daily\_Mean}_p + \beta_2\text{AQI}_t$.
		\ELSIF{$m =$ SARIMA}
		\STATE Define SARIMA$(p,d,q)(P,D,Q)_s$ on the univariate AQI series.
		\ELSIF{$m =$ MLP or MLP+Phys}
		\STATE Define a feed-forward MLP mapping $\mathbf{x}_t \mapsto \hat{y}_t$.
		\ELSIF{$m =$ LSTM or LSTM+Phys}
		\STATE Define a one-to-one LSTM-based model (Algorithm~3) mapping $[\text{Daily\_Mean}_p(t), \text{AQI}_t] \mapsto \hat{y}_t$.
		\ENDIF
		\ENDFOR
		
		\vspace{1mm}
		\STATE \textbf{Stage 3: Physics Guided AQI formulation} \STATE Define $\text{PhysicsAQI}(c_t, \mathcal{B})$ using the EPA piecewise linear AQI formula: $\text{AQI}_t^{\text{phys}} = (I_{\text{hi}} - I_{\text{lo}})\frac{(c_t - C_{\text{lo}})}{(C_{\text{hi}} - C_{\text{lo}})} + I_{\text{lo}}.$ \STATE Return $\text{AQI}_t^{\text{phys}}$ (scaled if required). \vspace{1mm}
		
		\vspace{1mm}
		\STATE \textbf{Stage 4: Training Loop for Each Model and Dataset}
		\FOR{each $(p, LAG)$}
		\FOR{each model type $m \in \mathcal{M}$}
		
		\IF{$m =$ SARIMA}
		\STATE Fit SARIMA on the raw training AQI series and skip gradient-based training.
		\ELSE
		\FOR{epoch $= 1$ to max\_epochs}
		\STATE Compute predictions on training data and evaluate data loss.
		\IF{$m$ is physics-informed}
		\STATE Compute physics-based AQI and physics loss.
		\STATE Total loss: $\mathcal{L} = \lambda_{\text{data}}\mathcal{L}_{\text{data}} + \lambda_{\text{phys}}\mathcal{L}_{\text{phys}}$.
		\ELSE
		\STATE $\mathcal{L} = \mathcal{L}_{\text{data}}$.
		\ENDIF
		\STATE Backpropagate $\mathcal{L}$ and update model parameters.
		\ENDFOR
		\ENDIF
		
		\ENDFOR
		\ENDFOR
		
		\vspace{1mm}
		\STATE \textbf{Stage 5: Testing and Evaluation}
		\FOR{each $(p, LAG)$ and model $m \in \mathcal{M}$}
		\STATE Generate predictions on train and test sets.
		\IF{$m \neq \mathrm{SARIMA}$}
		\STATE Apply inverse scaling to recover predictions and targets in real AQI units.
		\ENDIF
		\STATE Compute MAE, RMSE, and $R^2$ in real units and save in benchmarking table.
		\ENDFOR
		
	\end{algorithmic}
\end{algorithm}

The required dataset has been assembled and preprocessed using the procedure described in Algorithm \ref{alg:lag_aqi} in Section \ref{sec:dataset_description}. This pre-processed data feeds into a unified modeling stage where classical machine-learning models, deep-learning architectures, and physics-guided variants are applied to the same lag-wise datasets. The physics guidance incorporates EPA’s AQI formulation, which links pollutant concentration to AQI through breakpoint-based piecewise rules. Then, the structured workflow establishes a consistent, reproducible, and scientifically grounded foundation for evaluating AQI forecasting models.  By applying all the lag-wise datasets separately to several model families such as linear regression, SARIMA, MLP, and LSTM, and finally then the MLP+Physics and LSTM+Physics frameworks, a fair assessment of the role of model complexity and domain knowledge in predictive performance is made possible.The inclusion of physics-guided knowledge constraints ensures that the deep-learning models are not only optimized through hyperparameter tuning but also guided toward physically meaningful behavior.

\subsection{Baseline Statistical and Machine-Learning Models}
\subsubsection{Linear Regression}
One of the most commonly used statistical learning models is linear regression to represent the relationships between predictors and a continuous response variable, originating from the classical least-squares framework\cite{montgomery2021introduction, roustaei2024application}. As a baseline for the AQI forecasting, a linear regression model has been used and applied to the PM$_{2.5}$ and O$_3$ datasets among all the forecasting horizons with LAG 1, 7, 14, and 30, respectively. The model assumes a linear relationship between the current-day pollutant level and AQI and the next-day AQI target and provides an interpretable benchmark for comparing the performance of more advanced time-series and deep-learning models. For each dataset, the features consisted of the current day’s pollutant concentration and AQI value, while the target variable corresponded to the AQI value at the specified future horizon. To validate the effect of the normalization, both scaling and no scaling with StandardScaler are performed. When scaling was applied, the transformation parameters were learned exclusively from the training portion of the data and applied consistently to the test set to prevent information leakage. The linear regression model follows the standard Ordinary Least Squares formulation, where the relationship between inputs and the predicted AQI is modeled as

\begin{equation}
	\hat{y} = \beta_0 + \beta_1\,\text{Daily\_Mean\_Pollutant}(t)
	+ \beta_2\,\text{Daily\_AQI\_Value}(t),
\end{equation}

and the coefficients $\beta_0$, $\beta_1$, and $\beta_2$ are estimated by minimizing the residual sum of squares. This simple architecture provides a transparent, interpretable mapping from current-day conditions to the future AQI. This baseline allows a clear comparison between simple data-driven learning and more complex architectures such as SARIMA, MLP, LSTM, and the physics-guided PINN variants. 

\subsubsection{SARIMAX}
The Seasonal AutoRegressive Integrated Moving Average with eXogenous regressors (SARIMAX) is a model extending the classical Box-Jenkins ARIMA and SARIMA models \cite{box2015time} that adds external predictor(s) to the system. It permitted capturing both internal temporal structure and exogenous influences in time series forecasting and was fast-used in weather forecasting as well \cite{vagropoulos2016comparison, alharbi2022seasonal, lee2025performance}.  In order to focus on both short-term autoregressive nature and seasonal patterns from the datasets, the SARIMAX model is trained and tested on all lag-wise datasets, which will ensure the primary classical time series baseline for benchmarking. A fixed configuration with non-seasonal order and seasonal order is employed, which reflects both short-term AQI dynamics as well as weekly seasonal patterns for AQI and is applied the same way for all LAG 1, 7, 14, and 30 datasets and appropriately aligned before training. For each lag-wise dataset, the current day AQI series is taken as an endogenous variable, whereas the previous day's pollutant concentration is added as a lagged exogenous regressor. This enables the model to reflect the internal temporal characteristics of AQI and the external influence of pollutant fluctuation with time. StandardScaler is always fitted only on the training set of each dataset, to prevent information leakage before fitting the test set using the same parameters. The model is implemented using the Statsmodels SARIMAX framework and is fitted using maximum likelihood estimation by CPU resources because GPU acceleration is not providing extra benefits for this class of statistical models. Separate experiments are carried out for each forecasting horizon, which makes it possible to assess the variation in difficulty of prediction for the short- and medium-term horizons. \noindent{}The SARIMAX architecture combines the non-seasonal ARIMA components with a seasonal ARIMA structure and an exogenous pollutant regressor. In this setup, the AQI value at time $t$ is influenced by its own past values, past error terms, weekly seasonal dependencies, and the lagged pollutant concentration. The general model formulation is expressed as:
\begin{equation}
	y_t = \phi_1 y_{t-1} + \theta_1 \varepsilon_{t-1} + \Phi_1 y_{t-7} + \Theta_1 \varepsilon_{t-7} + \beta x_{t-1} + \varepsilon_t,
\end{equation}
where $y_t$ denotes the AQI at time $t$, $x_{t-1}$ is the pollutant concentration from the previous day, and $\varepsilon_t$ is the model residual. This formulation captures both short-term dependencies and recurring weekly seasonal effects in AQI while integrating pollutant-driven variations. This SARIMAX configuration therefore provides a consistent and interpretable classical baseline against which the performance of the MLP, LSTM, and physics-guided PINN models can be systematically compared.

\subsection{Deep-Learning Models for AQI Forecasting}

\subsubsection{Multilayer Perceptron (MLP) Model}

Multilayer Perceptrons (MLPs) \cite{popescu2009multilayer} is among the first known and very popular neural network architectures for nonlinear function approximation, and these were successfully applied in weather and environmental forecasting problems since they have the capacity to model complex relationships between atmospheric variables and future weather conditions \cite{abhishek2012weather, al2025data}. The MLP model is utilized as a nonlinear supervised baseline in between the simple linear regression and more complicated models (LSTM and physics-guided models). For each pollutant (PM$_{2.5}$ and O$_3$) and each forecasting horizon (LAG 1, 7, 14, and 30), the model is trained using the lag-wise datasets as mentioned earlier. In each case, the input features consist of the current day’s pollutant concentration and AQI value, while the target corresponds to the AQI value at the chosen future lag. The data are split chronologically using an 80\%–20\% train–test partition without shuffling to preserve temporal order. StandardScaler is applied to the input features, and these parameters are learned from the training part only, then reused to transform the test set, whereas the targeted AQI remains in its original value. The MLP design is built using PyTorch as a fully connected feedforward network with two input neurons (for pollutant and AQI) and two hidden layers with 64 and 32 neurons, respectively. Each hidden layer uses ReLU activation to capture nonlinear interactions. The forward propagation of the MLP can be expressed as:

\begin{equation}
	\hat{y}_{t+L}  
	= W_{3}\,\sigma\!\left( W_{2}\,\sigma\!\left( W_{1} x_{t} + b_{1} \right) + b_{2} \right) + b_{3},
\end{equation}

where $x_t = [\text{PM$_{2.5}$}(t)/O_3(t),\, \text{AQI}(t)]^\top$ is the two-dimensional input vector, $W_1$, $W_2$, and $W_3$ denote the learnable weight matrices, $b_1$, $b_2$, and $b_3$ are the corresponding biases, and $\sigma(\cdot)$ represents the ReLU activation function. A last linear output neuron makes a single scalar prediction for the AQI value in the future. The AdamW optimizer with a learning rate of 0.001 and a mean squared error (MSE) loss calculated directly in AQI units is used to train the model across a set number of epochs. MAE, RMSE, and R² evaluation metrics are used to measure the model's performance on a held-out test set after training. Such MLP model provided us a data-driven benchmark which is nonlinear and that can be used to compare with the other ML/DL models and their physics-informed versions.

\subsubsection{Long Short-Term Memory (LSTM) Model}

Long Short-Term Memory (LSTM) \cite{hochreiter1997long} works on sequential data and is very effective for predicting long-range time series dependencies, which use RNN to memorize the information from earlier events and make decisions accordingly \cite{graves2012long}. LSTM is used in weather forecasting and air quality data analysis for its effective decision-making on time-series data \cite{yan2021multi, zhang2025machine}. A LSTM network has been developed as a nonlinear sequence-based baseline to capture temporal dependencies in AQI evolution. For each pollutant (PM$_{2.5}$ and O$_3$) and forecasting horizon (LAG 1, 7, 14, and 30), the model is trained on the same lag-wise datasets used for the MLP and statistical baselines. Each lag-wise dataset provides two input features: the current day’s pollutant concentration and AQI value along with the corresponding future AQI value at the selected forecasting horizon. An 80\%--20\% chronological split is used to preserve forecasting realism. To ensure consistent preprocessing, min-max normalization in the range $[-1, 1]$ is fitted only on the training portion of each dataset using all three relevant columns (pollutant, AQI, and future AQI). The same transformation is then applied to the test set, preventing any information leakage. For the model architecture, the sequence length is settled to $T=1$ along with two features, $F=2$, that produced tensors of shape that are perfect for building an LSTM model.

The model is built with PyTorch, which consists of two stacked unidirectional LSTM layers that have 32 hidden units individually and are then followed by deep, fully connected layers of 128, 64, and 32 neurons with activation functions and dropout regularization. The forward computation of each LSTM cell follows the standard gated formulation:

\begin{align}
	\begin{aligned}
		i_t &= \sigma(W_i x_t + U_i h_{t-1} + b_i),\\
		f_t &= \sigma(W_f x_t + U_f h_{t-1} + b_f),\\
		o_t &= \sigma(W_o x_t + U_o h_{t-1} + b_o),\\
		\tilde{c}_t &= \tanh(W_c x_t + U_c h_{t-1} + b_c),\\
		c_t &= f_t \odot c_{t-1} + i_t \odot \tilde{c}_t,\\
		h_t &= o_t \odot \tanh(c_t),
	\end{aligned}
\end{align}

where $x_t \in \mathbb{R}^2$ contains the current pollutant level and AQI value, $i_t$, $f_t$, and $o_t$ denote the input, forget, and output gates, $c_t$ is the cell state, and $h_t$ is the hidden state passed to subsequent layers. Then a final linear neuron provided the final output in scaled units. The model is optimized by using Adam and a learning rate of 0.0001, followed by MSE loss. The learning rate is adjusted by a ReduceLROnPlateau scheduler when validation loss plateaus, and we used early stopping with a patience of 20 epochs to ensure the generalization. After training, predictions are inverse-transformed back to real AQI units, and evaluation is performed using R$^2$, MAE, and RMSE. This LSTM baseline provides a sequence-oriented deep-learning benchmark for comparison with the MLP and the physics-informed LSTM variants.

\subsection{Proposed Physics-Informed Model for AQI Prediction}

A physics-guided machine learning model helps the model to learn more about the domain knowledge and incorporate it into the overall learning process \cite{raissi2019physics}. To analyze and benchmark the knowledge guidance on AQI prediction, a physics-guided model is proposed in this research. We implemented and applied the domain knowledge of the AQI calculation directly to the ML/DL model and proposed one hybrid model with MLP+Physics and another hybrid deep learning model, LSTM+Physics. Domain knowledge information for calculating the AQI equation provided by the U.S. EPA has been applied to our developed neural network \cite{EPA_AirData_Daily}. It is worth mentioning that our baseline models (LR, SARIMAX, MLP, and LSTM) are completely dependent on the statistical data and the relationship between PM and ozone concentrations, as well as on future AQI values. However, in this research, the proposed MLP+Physics and LSTM+Physics models provided the domain knowledge for AQI calculation that powered the learning function to stay more consistent. The AQI is defined using a piecewise linear function, in which each pollutant concentration interval \([C_{\text{low}}^{(k)},\, C_{\text{high}}^{(k)}]\) corresponds to an AQI interval \([I_{\text{low}}^{(k)},\, I_{\text{high}}^{(k)}]\). For any concentration value \(C_i\) that falls within the \(k\)-th EPA breakpoint range, the AQI is computed as:

\begin{equation}
	f_{\text{AQI}}(C_i)
	= I_{\text{low}}^{(k)}
	+ \frac{I_{\text{high}}^{(k)} - I_{\text{low}}^{(k)}}{C_{\text{high}}^{(k)} - C_{\text{low}}^{(k)}}
	\left(C_i - C_{\text{low}}^{(k)}\right),
	\qquad C_i \in [C_{\text{low}}^{(k)}, C_{\text{high}}^{(k)}].
	\label{eq:aqi_piecewise}
\end{equation}

This formulation is applied using the full EPA \cite{epa2024_aqi_tech_doc} breakpoint tables for both PM\(_{2.5}\) and O\(_3\), covering all AQI categories from ``Good'' to ``Hazardous,'' and the computed AQI values are capped at the EPA maximum of 500. The output of this function serves as a physics-based AQI reference that guides the neural models toward pollutant--AQI relationships that adhere to established atmospheric behavior.

To embed this physical knowledge into the learning process, the supervised data loss is combined with a physics-based consistency loss. The standard data-driven loss is defined as:
\begin{equation}
	\mathcal{L}_{\text{data}}
	= \frac{1}{N}\sum_{i=1}^{N} (\hat{y}_i - y_i)^2,
	\label{eq:data_loss}
\end{equation}
where \(\hat{y}_i\) is the model prediction and \(y_i\) is the true AQI value.

The physics loss penalizes deviations between the model prediction and the EPA-derived AQI computed from pollutant concentrations:
\begin{equation}
	\mathcal{L}_{\text{phys}}
	= \frac{1}{N}\sum_{i=1}^{N} \left( \hat{y}_i - f_{\text{AQI}}(C_i) \right)^2.
	\label{eq:phys_loss}
\end{equation}

The total loss combines these two terms using a weighted formulation:
\begin{equation}
	\mathcal{L}_{\text{total}}
	= \lambda_{\text{data}} \, \mathcal{L}_{\text{data}}
	+ \lambda_{\text{phys}} \, \mathcal{L}_{\text{phys}},
	\label{eq:total_loss}
\end{equation}

where \(\lambda_{\text{data}}\) and \(\lambda_{\text{phys}}\) control the balance between empirical data fitting and adherence to physically grounded AQI behavior. By adjusting these weights, the framework spans purely data-driven learning (\(\lambda_{\text{phys}} = 0\)) to fully physics-constrained modeling (\(\lambda_{\text{data}} = 0\)). This unified physics-informed formulation is applied consistently across both the MLP+Physics and LSTM+Physics architectures, enabling systematic evaluation of physics-guided deep learning for AQI forecasting across pollutants and multi-horizon settings.

\subsubsection{MLP+Physics}

The proposed MLP+Physics model incorporates the EPA AQI formulation \cite{epa2024_aqi_tech_doc} for both PM$_{2.5}$ and O$_3$, and each pollutant-specific model is trained separately following the same procedure used for the baseline MLP. The physics formula is imposed directly within the training loss function. The model architecture mirrors the baseline design: the input vector $x_t$ contains two features, the current day's pollutant concentration and AQI value that scaled using a \texttt{StandardScaler} fitted strictly on the training portion of each lag-wise dataset. The network contains two fully connected hidden layers with 64 and 32 neurons, respectively, each followed by a ReLU activation, and a final linear output neuron that produces a single predicted AQI value in real AQI units.

During training, the forward pass produces the model prediction $\hat{y}_i$, while the pollutant concentration $C_i$ is simultaneously passed through the EPA breakpoint-based AQI computation to generate the physics-derived reference value $f_{\text{AQI}}(C_i)$ defined in Equation~\eqref{eq:aqi_piecewise}. The data-driven loss follows the form given in Equation~\eqref{eq:data_loss}, and the physics-based loss follows Equation~\eqref{eq:phys_loss}. These two components are combined using scalar weights $\lambda_{\text{data}}$ and $\lambda_{\text{phys}}$ as defined in the total loss expression in Equation~\eqref{eq:total_loss}. Optimization is performed using the AdamW optimizer with a learning rate of 0.001 over a fixed number of epochs. This hybrid design enforces simultaneous learning from empirical data and established pollutant--AQI relationships. Because the MLP architecture remains constant across horizons, this physics-guided model provides a consistent nonlinear benchmark for assessing the influence of domain knowledge on prediction performance across pollutants and forecasting horizons.

\begin{figure*}[htbp]
	\centering
	\includegraphics[scale=0.35]{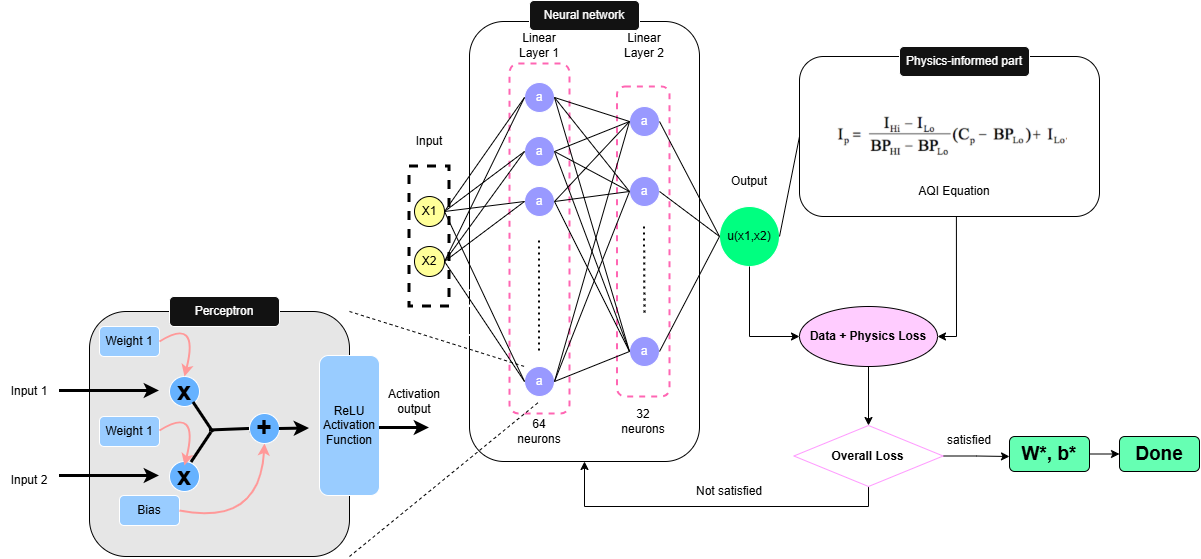}
	\caption{
		Architecture of the MLP+Physics Model.
	}
	\label{fig:MLP+Physics}
\end{figure*}

\subsubsection{LSTM + Physics Integration}

The proposed LSTM+Physics model has been implemented by using the same baseline LSTM model architecture and integrating Equation~\eqref{eq:aqi_piecewise} into the learning part of the model. This model is trained individually for PM\(_{2.5}\) and O\(_3\) with the lag-wise datasets of the multi-horizons. The input sequence is designed by the current day's pollutant concentration along with the AQI value. The target variable is defined as the AQI from the future lag days within the forecasting horizon. The sequence data is normalized using a Min--Max scaler fitted strictly on the training portion of each dataset. First of all, two stacked unidirectional LSTM layers have been used with 32 hidden layers for each, and a multilayer fully connected layer is being followed with 128, 64, and 32 neurons and activated by ReLU, and then dropout regularization is applied. Finally, a linear neuron outputs the predicted AQI value in the scaled domain that is inverse-transformed later to real AQI units for evaluation.

To train the model effectively, a forward pass created the prediction $\hat{y}_i$ of the model for every sample input, and the respective pollutant concentration $C_i$ is than provided to the AQI function $f_{\text{AQI}}(C_i)$ defined in Equation~\eqref{eq:aqi_piecewise}, that helps and provides the domain knowledge to the training process. Data-driven loss has been calculated by using the MSE formula described in Equation~\eqref{eq:data_loss}, and the physics loss has been calculated from Equation~\eqref{eq:phys_loss}. Finally, total loss has been calculated from the data loss and physics loss by the equation~\eqref{eq:total_loss}. Overall, this process ensures a balance between learning and physical consistency in order to accurately detect the AQI based on pollutant concentration.

The model is optimized using the Adam optimizer, which has a learning rate of 0.0001. A ReduceLROnPlateau scheduler adjusts the learning rate when the validation loss reaches saturation, and early pausing with a 20-epoch patience is employed to promote generalization. By including the physics-based AQI constraints into the sequence-learning architecture, the LSTM+Physics model enforces domain-aware temporal reasoning. This strengthens the benchmark for assessing how physical knowledge enhances AQI forecasting over many horizons and contaminants.

\begin{figure*}[htbp]
	\centering
	\includegraphics[scale=0.35]{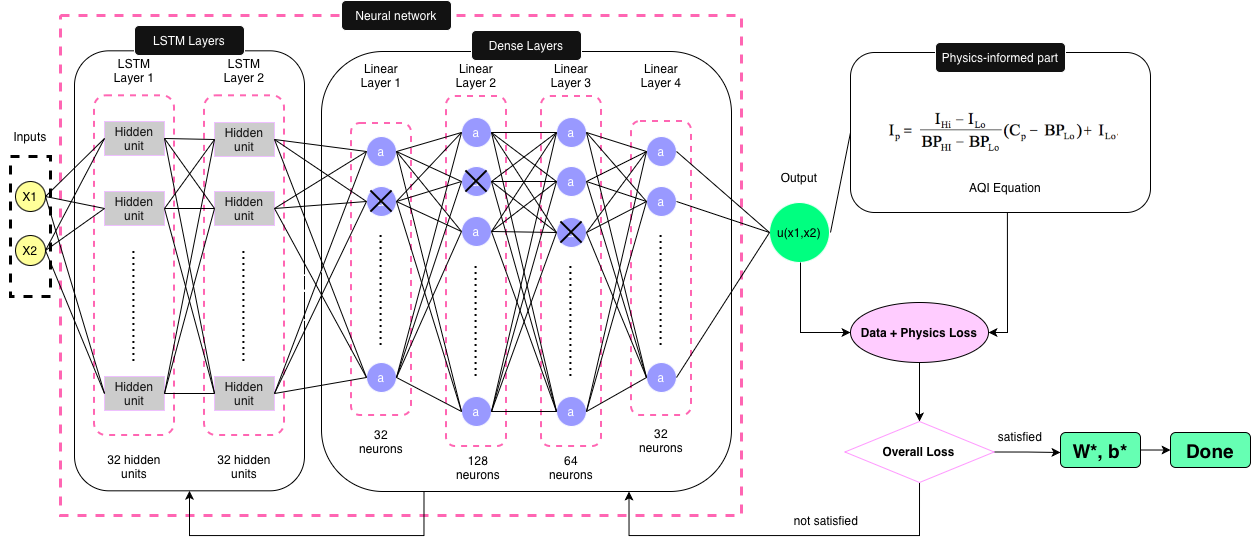}
	\caption{
		Architecture of the LSTM+Physics Model.
	}
	\label{fig:LSTM+Physics}
\end{figure*}

\section{Experimental Setup and Evaluation Metrics}
A unified experimental protocol was adopted to ensure fair and reproducible comparison across classical, machine learning, deep learning, and physics-guided models for multi-horizon AQI forecasting.
\subsection{Hardware and Software Specifications}
A personal CPU-based computing system with an Intel Core i5 processor and 8 GB of RAM is used for the experiment of this research study and all the models were implemented in PyTorch. The software environment was based on Python, with data preprocessing and classical models implemented using NumPy, Pandas, scikit-learn, and statsmodels. Neural network–based and physics-guided models were implemented using PyTorch. To ensure reproducibility, fixed random seeds were applied across all experiments, and all datasets were split chronologically to avoid temporal data leakage.

\subsection{Model Hyperparameter Configuration}
A concise summary of the hyperparameter settings is showen in the Table \ref{tab:hyperparameters} which is used in the evaluated models. It represents the common arcitectural and optmization paramteres that are used in the experiments. Classical models, including Linear Regression and SARIMAX, rely on fixed analytical formulations and therefore do not require iterative optimization or feature scaling. 
\begin{table}[htbp]
	\centering
	\caption{Generalized hyperparameter settings for all evaluated models}
	\label{tab:hyperparameters}
	\renewcommand{\arraystretch}{1.2}
	\resizebox{\textwidth}{!}{%
		\begin{tabular}{lcccccc}
			\hline
			\textbf{Hyperparameter} 
			& \textbf{LR} 
			& \textbf{SARIMAX} 
			& \textbf{MLP} 
			& \textbf{MLP+Physics} 
			& \textbf{LSTM} 
			& \textbf{LSTM+Physics} \\
			\hline
			
			Architecture / Order 
			& OLS 
			& (1,1,1)$\times$(1,0,1,7) 
			& 64--32 
			& 64--32 
			& LSTM(32,32)+MLP 
			& LSTM(32,32)+MLP \\
			
			Activation 
			& -- 
			& -- 
			& ReLU 
			& ReLU 
			& ReLU 
			& ReLU \\
			
			Optimizer 
			& -- 
			& -- 
			& AdamW 
			& Adam 
			& Adam 
			& Adam \\
			
			Learning rate 
			& -- 
			& -- 
			& 0.001 
			& 0.001 
			& 0.001 
			& 0.001 \\
			
			Batch size 
			& -- 
			& -- 
			& Full 
			& Full 
			& 32 
			& 32 \\
			
			Scaling method 
			& None 
			& None 
			& StandardScaler 
			& StandardScaler 
			& MinMaxScaler [$-1,1$] 
			& MinMaxScaler [$-1,1$] \\
			
			\hline
		\end{tabular}
	}
\end{table}

\subsection{Evaluation Metrics}

To evaluate the forecasting performance of the proposed models, standard regression-based error metrics are used. These metrics quantify the difference between the predicted AQI values and the observed (true) AQI values on the test dataset. All evaluation metrics are computed in real AQI units to ensure clear interpretation and practical relevance.

\subsubsection{Mean Absolute Error (MAE)}

Mean Absolute Error (MAE) measures the average magnitude of the prediction errors without considering their direction. It indicates how far, on average, the predicted AQI values deviate from the observed values and is less sensitive to large outliers.

\begin{equation}
	\text{MAE} = \frac{1}{N} \sum_{i=1}^{N} \left| y_i - \hat{y}_i \right|
\end{equation}

\subsubsection{Mean Squared Error (MSE)}

Mean Squared Error (MSE) computes the average of the squared differences between the predicted and observed AQI values. By squaring the errors, MSE penalizes larger prediction errors more heavily, making it useful for assessing model stability and robustness.

\begin{equation}
	\text{MSE} = \frac{1}{N} \sum_{i=1}^{N} \left( y_i - \hat{y}_i \right)^2
\end{equation}

\subsubsection{Normalized Mean Squared Error (NMSE)}

Normalized Mean Squared Error (NMSE) is used to evaluate the prediction error relative to the variability of the observed AQI values. By normalizing the MSE with respect to the variance of the ground-truth data, NMSE allows fair comparison across different pollutants and forecasting horizons. Lower NMSE values indicate better predictive performance, while values close to or greater than 1 suggest poor model accuracy relative to the data variance.

\begin{equation}
	\text{NMSE} = \frac{\frac{1}{N} \sum_{i=1}^{N} \left( y_i - \hat{y}_i \right)^2}
	{\frac{1}{N} \sum_{i=1}^{N} \left( y_i - \bar{y} \right)^2}
\end{equation}

Here, $y_i$ denotes the observed AQI value, $\hat{y}_i$ is the predicted AQI value, $\bar{y}$ represents the mean of the observed AQI values, and $N$ is the total number of samples.

\section{Results \& Comparative Analysis}

This section presents the empirical performance of all of the forecasting models for two pollutants (PM$_{2.5}$ and O$_3$) and four forecasting horizons ($LAG \in \{1,7,14,30\}$ days). We first analyze the behavior of the multi-horizon AQI targets to understand the changing difficulty in forecasting based on the time gap. Next, we compare baseline statistical and deep-learning models (LR, SARIMAX, MLP, and LSTM). Finally, we evaluate physics-guided variants (MLP+Physics and LSTM+Physics) and discuss how the EPA breakpoint-based AQI constraint influences predictive accuracy and stability across pollutants and horizons.

\subsection{Descriptive Analysis of Multi-Horizon AQI Targets}
\begin{figure*}[h]
	\centering
	\includegraphics[width=\textwidth]{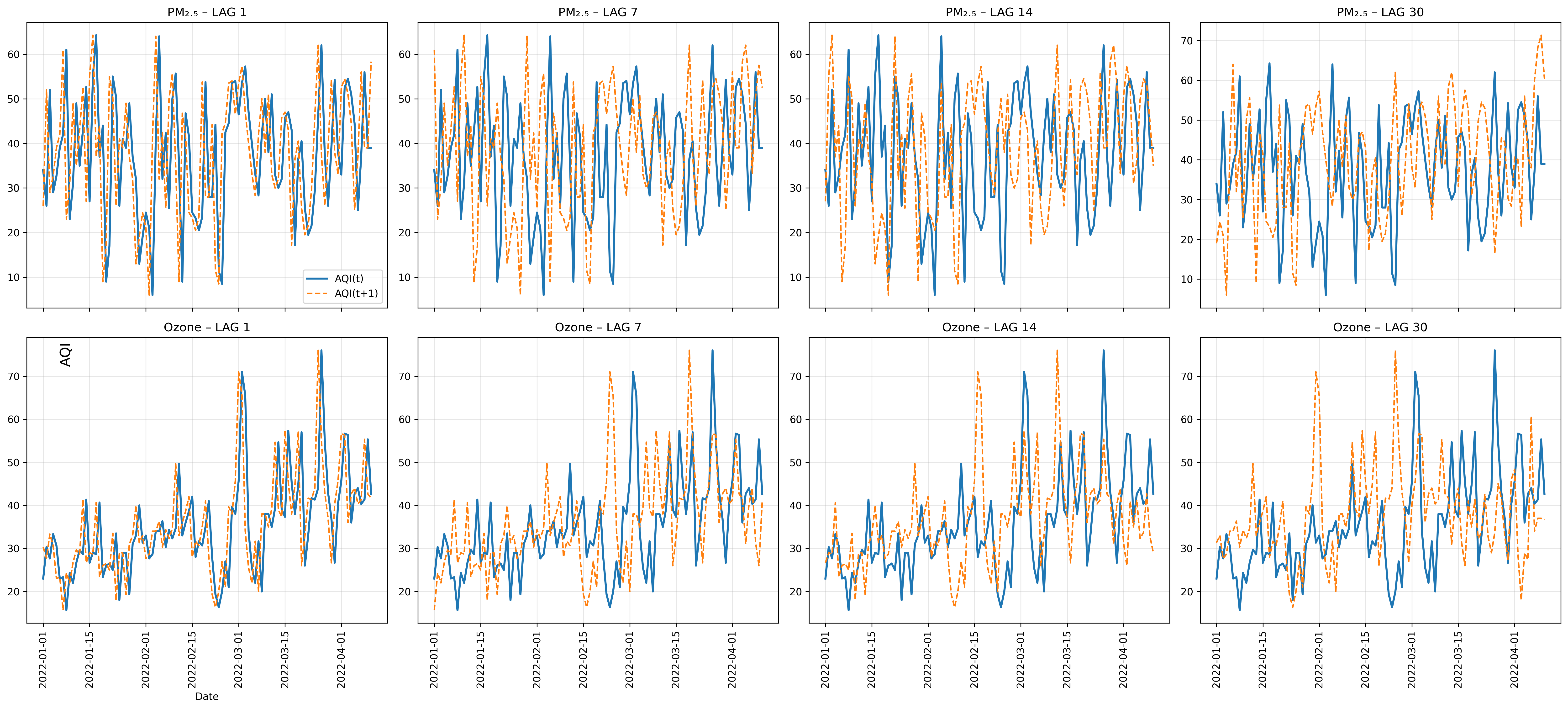}
	\caption{
		Multi-horizon AQI target visualization for PM$_{2.5}$ (top row) and O$_3$ (bottom row). 
		Each subplot compares the current-day AQI (t) with future AQI values at 1, 7, 14, and 30-day LAG's prediction.
	}
	\label{fig:pm_ozone_multihorizon}
\end{figure*}

Figure~\ref{fig:pm_ozone_multihorizon} shows the relation between the present AQI at time $t$ and the future goals at $t+1$, $t+7$, $t+14$, and $t+30$ in the case of PM$_{2.5}$ (top row) and O$_3$ (bottom row). With $LAG=1$, the goals are still rather near to the current AQI. This indicates that there is a whole lot of short-term persistence and the forecast is pretty stable. As the AQI forecasting horizons become closer to $LAG=7$, $14$, and $30$, the future AQI becomes less dependent on the present and more variable, which indicates that it is less dependent on time. As this disparity gets larger with respect to time, that is why prediction errors tend to go up.
 
For \textbf{PM$_{2.5}$}, the AQI targets exhibit smoother and more gradual temporal changes. PM$_{2.5}$ maintains relatively stable patterns across all horizons, indicating better predictability compared to O$_3$. However, relatively with it, O$_3$ have considerably higher daily swings, particularly over longer time horizons which implying that O$_3$ AQI forecasting is less stable and more challenging over longer time periods. Overall, the figure emphasizes the use of more expressive learning approaches (MLP/LSTM) and domain guidance (physics constraints) to boost resilience, as well as the use of multi-horizon forecasting models.

\subsection{Baseline Model Performance Across Lag-Wise Data}

We benchmark four baseline models such as linear regression (LR), SARIMAX, multilayer perceptron (MLP), and long short-term memory (LSTM) based on a lag-wise dataset of PM$_{2.5}$ and O$_3$. For each pollutant and horizon, we repored the distribution of MAE, RMSE, and NMSE on the held-out chronological test set. These baseline results provided a point of reference for before this physics guidance is introduced.

\subsubsection{\texorpdfstring{PM$_{2.5}$ Baseline Results}{PM2.5 Baseline Results}}

Table~\ref{tab:baseline_pm25_all} represents the overall baseline performance on PM$_{2.5}$. We found that, for $LAG=1$, the MLP and LSTM models achieve lower errors than simpler baselines, which indicates that nonlinear models better capture the pollutant AQI relationships. As when we increased the horizon to ($LAG=7$, $14$, and $30$), all models showed more or less efficiency, reflecting the increased uncertainty of longer-term forecasts. Among the baselines, LSTM is quite stable across horizons, which seems to indicate that sequence-based modeling performs better in generalization as the prediction gap becomes larger.

\begin{table*}[h]
	\centering
	\caption{Baseline performance of LR, SARIMAX, MLP, and LSTM models for PM$_{2.5}$ AQI forecasting across all lag-wise datasets. 
	}
	\label{tab:baseline_pm25_all}
	\vspace{2mm}
	
	\resizebox{\textwidth}{!}{
		\begin{tabular}{lcccccccccccc}
			\toprule
			\multirow{2}{*}{\textbf{Model}} 
			& \multicolumn{3}{c}{\textbf{LAG 1}}
			& \multicolumn{3}{c}{\textbf{LAG 7}} 
			& \multicolumn{3}{c}{\textbf{LAG 14}}
			& \multicolumn{3}{c}{\textbf{LAG 30}} \\
			
			\cmidrule(lr){2-4} 
			\cmidrule(lr){5-7} 
			\cmidrule(lr){8-10}
			\cmidrule(lr){11-13}
			
			& MAE & RMSE & NMSE
			& MAE & RMSE & NMSE
			& MAE & RMSE & NMSE
			& MAE & RMSE & NMSE \\
			
			\midrule
			LR &
			8.2803 & 10.7370 & 0.6646 &
			10.2012 & 13.1408 & 0.9986 &
			10.1016 & 12.6088 & 0.9894 & 
			10.2011 & 12.7831 & 1 \\
			SARIMAX   & 
			9.8861 & 12.9285 & 0.8967 &
			31.7073 & 33.8115 & 0.9994 &
			48.4128 & 50.1106 & 1 & 
			12.4783 & 15.2045 & 0.9999 \\
			
			MLP &
			8.2149 & 10.6535 & 0.6583 &
			10.2289 & 13.1867 & 1.0305 &
			10.0839 & 12.6538 & 1.0185 &
			10.2548 & 12.8299 & 1.0464 \\
			
			LSTM &
			8.1761 & 10.6289 & 0.6552 &
			10.0669 & 12.9638 & 0.9960 &
			9.9955 & 12.5401 & 1.0003 &
			9.9895 & 12.5433 & 1.0002 \\
						
			\bottomrule
		\end{tabular}
	}
\end{table*}

\subsubsection{\texorpdfstring{O$_3$ Baseline Results}{O3 Baseline Results}}

Results for O$_3$ at the baseline appear in Table~\ref{tab:baseline_o3_all}. Similar to PM$_{2.5}$, the smallest errors appear at $LAG=1$ and the prediction errors grow as the horizons grow larger. MLP and LSTM roughly give better accuracy than simpler baselines, in line with their capacity to learn nonlinear patterns. Compared with PM$_{2.5}$, O$_3$ has more variation (Figure~\ref{fig:pm_ozone_multihorizon}) and less performance in long horizons correspondingly. Overall, these results indicate that O$_3$ is more difficult to accurately predict for relatively long periods.

\begin{table*}[h]
	\centering
	\caption{Baseline performance of LR, SARIMAX, MLP, and LSTM models for O$_3$ AQI forecasting across all lag-wise datasets. 
		}
	\label{tab:baseline_o3_all}
	\vspace{2mm}
	
	\resizebox{\textwidth}{!}{
		\begin{tabular}{lcccccccccccc}
			\toprule
			\multirow{2}{*}{\textbf{Model}} 
			& \multicolumn{3}{c}{\textbf{LAG 1}}
			& \multicolumn{3}{c}{\textbf{LAG 7}} 
			& \multicolumn{3}{c}{\textbf{LAG 14}}
			& \multicolumn{3}{c}{\textbf{LAG 30}} \\
			
			\cmidrule(lr){2-4} 
			\cmidrule(lr){5-7} 
			\cmidrule(lr){8-10}
			\cmidrule(lr){11-13}
			
			& MAE & RMSE & NMSE
			& MAE & RMSE & NMSE
			& MAE & RMSE & NMSE
			& MAE & RMSE & NMSE \\
			
			\midrule
			LR & 
			12.8202 & 20.2004 & 0.6357 &
			17.1723 & 25.9813 & 1 &
			16.1777 & 24.0898 & 0.8936 & 
			17.7215 & 25.5043 & 0.9892 \\
			
			SARIMAX   & 
			15.5971 & 22.6346 & 0.8065 &
			16.997 & 25.4225 & 0.9910 &
			17.8621 & 25.4614 & 1 & 
			29.7733 & 33.3162 & 1 \\
			
			MLP &
			12.8785 & 20.2860 & 0.6451 &
			17.4830 & 26.4023 & 1.0877 &
			16.0078 & 23.8830 & 0.8860 &
			17.4687 & 25.2393 & 0.9768 \\
			LSTM &
			12.7642 & 19.4974 & 0.5959 &
			16.7570 & 25.2099 & 0.9917 &
			16.3613 & 23.9691 & 0.8924 &
			17.7549 & 25.3641 & 0.9865 \\
			\bottomrule
		\end{tabular}
	}
\end{table*}

\subsection{Proposed Physics-Guided Model Performance Across Lag-Wise Data}

The proposed physics-guided models (MLP+Physics and LSTM+Physics) are evaluated, which use the EPA breakpoint-based AQI mapping as a consistency constraint during the training. The goal would be to help increase predictive stability and allow for the physical meaning of pollutant-AQI relationships. We compare the results of physics-guided with the corresponding baseline models using MAE, RMSE, and NMSE for all horizons.

\subsubsection{MLP+Physics Results}

Multiple results from MLP+Physics obtained using different settings of loss weights $(\lambda_{\text{data}},  \lambda_{\text{phys}})$ are shown in frequencies in Table~\ref{tab:pm25_physics_results} and Table~\ref{tab:O3_physics_results}. For PM${2.5}$ (Table~\ref{tab:pm25_physics_results}) physics guidance is suggested to achieve small but consistent improvement over the baseline MLP for the majority of the horizons, with mainly significant improvement observed for the short horizon ($LAG=1$). In addition, the use of balanced weighting (e.g., $\lambda_{\text{data}}=0.5$, $\lambda_{\text{phys}}=0.5$) may improve the performance at medium horizons; this would suggest that moderate guidance from physics may aid generalization in situations where the forecast gap becomes greater.

\begin{table*}[h]
	\centering
	\caption{
		Physics-guided MLP performance for PM$_{2.5}$ across forecasting horizons
		with different $\lambda_{\text{data}}$ and $\lambda_{\text{phys}}$ settings.
	}
	\label{tab:pm25_physics_results}
	\vspace{2mm}
	
	\resizebox{\textwidth}{!}{
		\begin{tabular}{lccccccccccccccc}
			\toprule
			\textbf{Model} &
			$\lambda_{\text{data}}$ &
			$\lambda_{\text{phys}}$ &
			\multicolumn{3}{c}{\textbf{LAG 1}} &
			\multicolumn{3}{c}{\textbf{LAG 7}} &
			\multicolumn{3}{c}{\textbf{LAG 14}} &
			\multicolumn{3}{c}{\textbf{LAG 30}} \\
			
			\cmidrule(lr){4-6}
			\cmidrule(lr){7-9}
			\cmidrule(lr){10-12}
			\cmidrule(lr){13-15}
			
			& & &
			MAE & RMSE & NMSE &
			MAE & RMSE & NMSE &
			MAE & RMSE & NMSE &
			MAE & RMSE & NMSE \\
			
			\midrule
			MLP+Physics & 0.0 & 1.0 &
			\textbf{8.1695} & \textbf{10.6349} & \textbf{0.6560} &
			\textbf{10.1783} & \textbf{13.1374} & \textbf{1.0228} &
			10.0925 & 12.6804 & 1.0228 &
			\textbf{10.2176} & \textbf{12.7926} & \textbf{1.0404} \\
			
			MLP+Physics & 0.3 & 0.7 &
			8.1826 & 10.6395 & 0.6565 &
			10.1934 & 13.1508 & 1.0249 &
			10.1092 & 12.6937 & 1.0249 &
			10.2260 & 12.8006 & 1.0417 \\
			
			MLP+Physics & 0.5 & 0.5 &
			8.1918 & 10.6430 & 0.6570 &
			10.2015 & 13.1591 & 1.0262 &
			\textbf{10.0649} & \textbf{12.6396} & \textbf{1.0162} &
			10.2332 & 12.8081 & 1.0429 \\
			
			MLP+Physics & 0.7 & 0.3 &
			8.2017 & 10.6477 & 0.6576 &
			10.2152 & 13.1725 & 1.0283 &
			10.0713 & 12.6445 & 1.0170 &
			10.2399 & 12.8143 & 1.0439 \\
			
			MLP & 1.0 & 0.0 &
			8.2149 & 10.6535 & 0.6583 &
			10.2289 & 13.1867 & 1.0305 &
			10.0839 & 12.6538 & 1.0185 &
			10.2548 & 12.8299 & 1.0464 \\
			
			\bottomrule
		\end{tabular}
	}
\end{table*}

\begin{figure*}[htbp]
	\centering
	
	\begin{subfigure}[t]{0.9\textwidth}
		\centering
		\includegraphics[width=\textwidth]{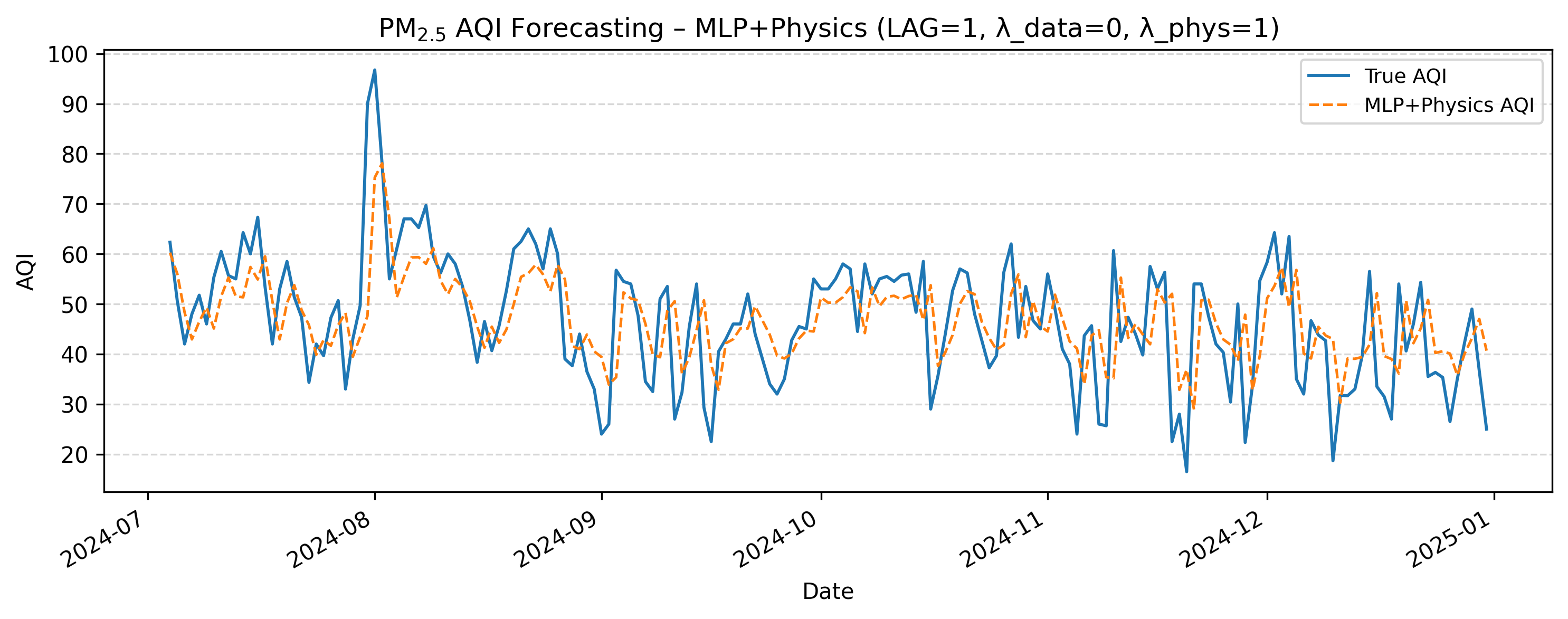}
		\caption{PM$_{2.5}$ AQI forecasting using MLP+Physics on the test subset.}
		\label{fig:pm25_mlp_physics_ts}
	\end{subfigure}
	
	\vspace{3mm}
	
	\begin{subfigure}[t]{0.9\textwidth}
		\centering
		\includegraphics[width=\textwidth]{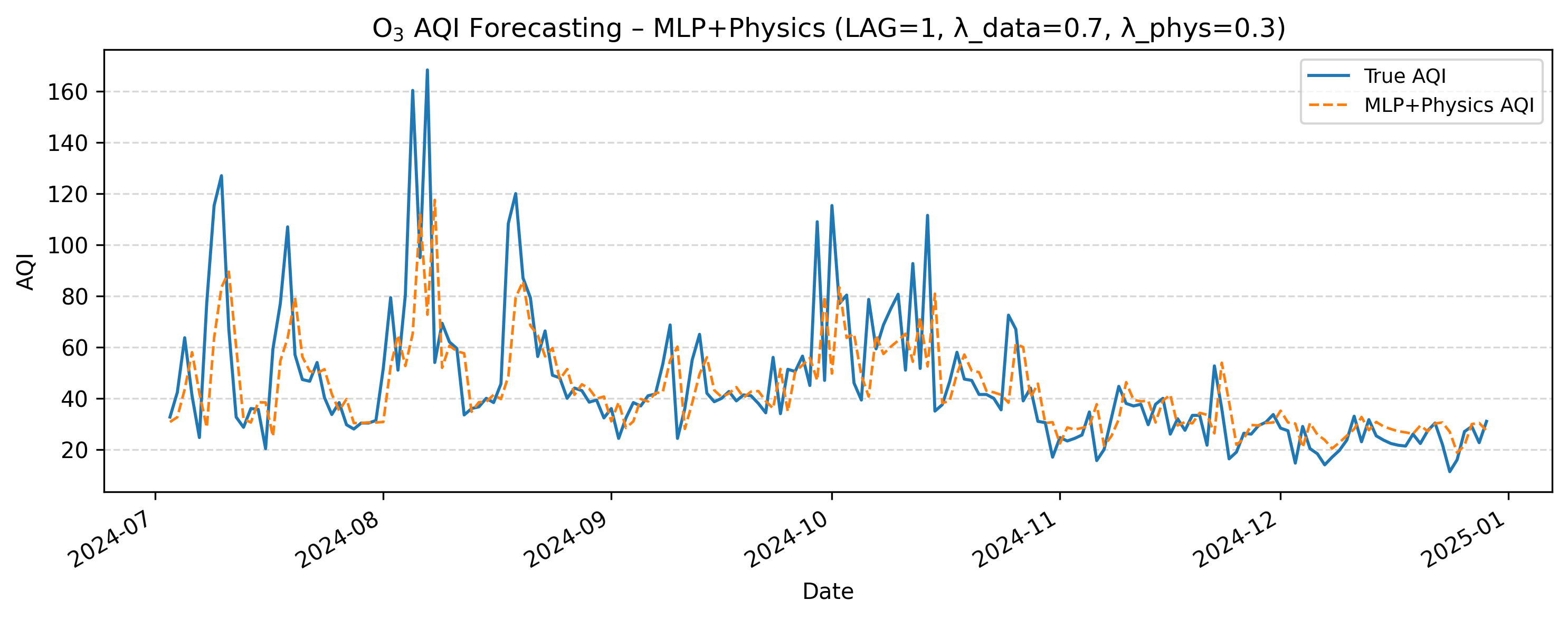}
		\caption{O$_3$ AQI forecasting using MLP+Physics on the test subset.}
		\label{fig:o3_mlp_physics_ts}
	\end{subfigure}
	
	\caption{
		Time-series comparison of true and predicted AQI values on the test set using the
		physics-guided MLP model for (a) PM$_{2.5}$ and (b) O$_3$.
		The plots show the last portion of the test period to highlight short-term forecasting behavior.
	}
	\label{fig:mlp_physics_timeseries}
\end{figure*}

For O$_3$ (Table~\ref{tab:O3_physics_results}), improvements are more modest and in several cases comparable to the baseline MLP. This indicates that the breakpoint-based constraint may provide weaker direct benefit for O$_3$ under daily aggregated observations, where AQI variability is higher. Figure~\ref{fig:mlp_physics_timeseries} provides qualitative comparisons of predicted and true AQI on the test subset for both pollutants, showing that physics guidance can improve stability and reduce unrealistic fluctuations in short-term forecasting behavior.

\begin{table*}[h]
	\centering
	\caption{
		Physics-guided MLP performance for O$_3$ across forecasting horizons
		with different $\lambda_{\text{data}}$ and $\lambda_{\text{phys}}$ settings.
	}
	\label{tab:O3_physics_results}
	\vspace{2mm}
	
	\resizebox{\textwidth}{!}{
		\begin{tabular}{lccccccccccccccc}
			\toprule
			\textbf{Model} &
			$\lambda_{\text{data}}$ &
			$\lambda_{\text{phys}}$ &
			\multicolumn{3}{c}{\textbf{LAG 1}} &
			\multicolumn{3}{c}{\textbf{LAG 7}} &
			\multicolumn{3}{c}{\textbf{LAG 14}} &
			\multicolumn{3}{c}{\textbf{LAG 30}} \\
			
			\cmidrule(lr){4-6}
			\cmidrule(lr){7-9}
			\cmidrule(lr){10-12}
			\cmidrule(lr){13-15}
			
			& & &
			MAE & RMSE & NMSE &
			MAE & RMSE & NMSE &
			MAE & RMSE & NMSE &
			MAE & RMSE & NMSE \\
			
			\midrule
			MLP+Physics & 0.0 & 1.0 &
			12.9227 & 20.2878 & 0.6452 &
			17.5359 & 26.4362 & 1.0905 &
			16.0406 & \textbf{23.8743} & \textbf{0.8853} &
			17.5058 & \textbf{25.2256} & \textbf{0.9757} \\
			
			MLP+Physics & 0.3 & 0.7 &
			12.9086 & 20.2862 & 0.6451 &
			17.5197 & 26.4273 & 1.0898 &
			16.0338 & 23.8792 & 0.8857 &
			17.4972 & 25.2312 & 0.9762 \\
			
			MLP+Physics & 0.5 & 0.5 &
			12.9000 & 20.2855 & 0.6451 &
			17.5082 & 26.4238 & 1.0895 &
			16.0278 & 23.8808 & 0.8858 &
			17.4954 & 25.2322 & 0.9763 \\
			
			MLP+Physics & 0.7 & 0.3 &
			12.8913 & \textbf{20.2854} & \textbf{0.6451} &
			17.4972 & 26.4139 & 1.0887 &
			16.0212 & 23.8823 & 0.8859 &
			17.4841 & 25.2338 & 0.9764 \\
			
			MLP & 1.0 & 0.0 &
			\textbf{12.8785} & 20.2860 & 0.6451 &
			\textbf{17.4830} & \textbf{26.4023} & \textbf{1.0877} &
			\textbf{16.0078} & 23.8830 & 0.8860 &
			\textbf{17.4687} & 25.2393 & 0.9768 \\
			
			\bottomrule
		\end{tabular}
	}
\end{table*}

\subsubsection{LSTM+Physics Results} 
Tables~\ref{tab:pm25_lstm_physics_results} and~\ref{tab:O3_lstm_physics_results} show the performance of the LSTM+Physics model across all forecasting horizons and loss-weight configurations for PM$_{2.5}$ and O$_3$, respectively. 

\begin{figure}[htbp]
	\centering
	
	\begin{subfigure}[t]{\columnwidth}
		\centering
		\includegraphics[width=\textwidth]{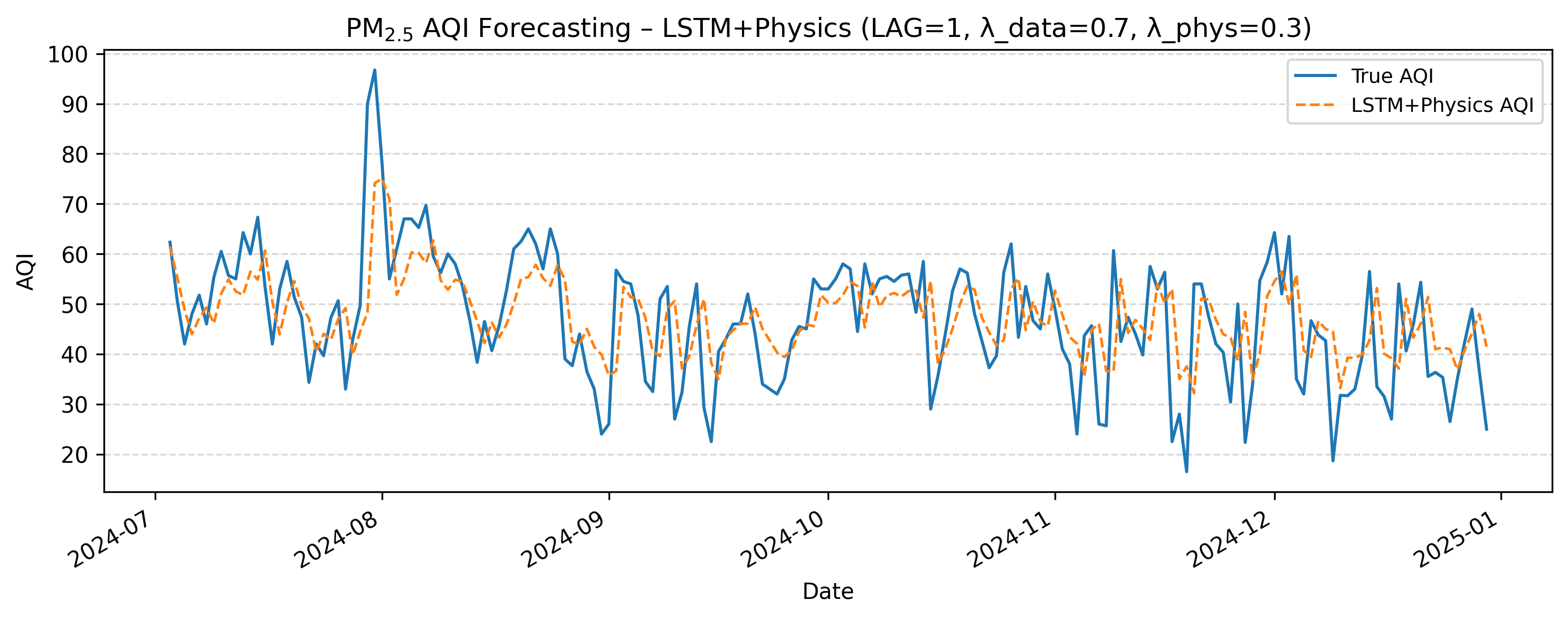}
		\caption{PM$_{2.5}$ AQI forecasting using LSTM+Physics on the test subset.}
		\label{fig:pm25_lstm_physics_ts}
	\end{subfigure}
	
	\vspace{2mm}
	
	\begin{subfigure}[t]{\columnwidth}
		\centering
		\includegraphics[width=\textwidth]{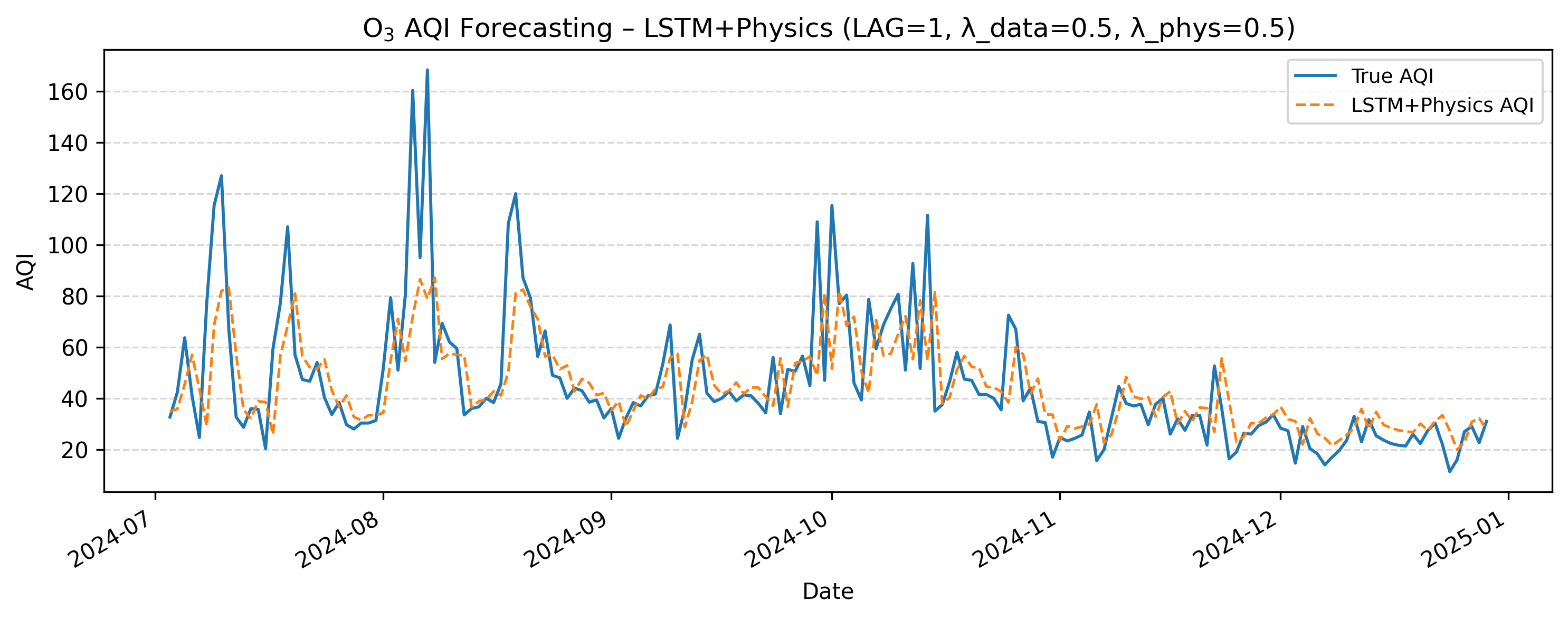}
		\caption{O$_3$ AQI forecasting using LSTM+Physics on the test subset.}
		\label{fig:o3_lstm_physics_ts}
	\end{subfigure}
	
	\caption{
		Time-series comparison of true and predicted AQI values on the test set using the
		physics-guided LSTM model for (a) PM$_{2.5}$ and (b) O$_3$.
	}
	\label{fig:lstm_physics_timeseries}
\end{figure}

PM$_{2.5}$ (Table~\ref{tab:pm25_lstm_physics_results}), the proposed LSTM+Physics model represents a strong and more stable performance across all over the AQI forecasting horizons. The optimal balance between data loss and physics loss varies slightly with the forecasting horizon, although the overall differences among weight settings remain small. Such results suggests that the LSTM sequence is more perfect and able to capture most of the temporal dynamics, while the physics-based constraint primarily contributes to improving physical consistency and stabilizing predictions rather than drastically reducing error values. This behavior is visually supported by Figure~\ref{fig:lstm_physics_timeseries}(a), where the predicted AQI closely follows the observed PM$_{2.5}$ AQI trajectory on the test subset.

For O$_3$ (Table~\ref{tab:O3_lstm_physics_results}), the baseline LSTM remains highly competitive, and the inclusion of physics guidance does not consistently lead to lower numerical errors across all horizons. Nevertheless, physics-informed training still plays an important role by enforcing physically meaningful pollutant--AQI relationships. As illustrated in Figure~\ref{fig:lstm_physics_timeseries}(b), the LSTM+Physics model tracks the observed O$_3$ AQI trends well over time and avoids unrealistic fluctuations, indicating improved temporal stability even when quantitative gains are modest.

\begin{table*}[htbp]
	\centering
	\caption{
		Physics-guided LSTM performance for PM$_{2.5}$ across forecasting horizons
		with different $\lambda_{\text{data}}$ and $\lambda_{\text{phys}}$ settings.
	}
	\label{tab:pm25_lstm_physics_results}
	\vspace{2mm}
	
	\resizebox{\textwidth}{!}{
		\begin{tabular}{lccccccccccccccc}
			\toprule
			\textbf{Model} &
			$\lambda_{\text{data}}$ &
			$\lambda_{\text{phys}}$ &
			\multicolumn{3}{c}{\textbf{LAG 1}} &
			\multicolumn{3}{c}{\textbf{LAG 7}} &
			\multicolumn{3}{c}{\textbf{LAG 14}} &
			\multicolumn{3}{c}{\textbf{LAG 30}} \\
			
			\cmidrule(lr){4-6}
			\cmidrule(lr){7-9}
			\cmidrule(lr){10-12}
			\cmidrule(lr){13-15}
			
			& & &
			MAE & RMSE & NMSE &
			MAE & RMSE & NMSE &
			MAE & RMSE & NMSE &
			MAE & RMSE & NMSE \\
			
			\midrule
			LSTM+Physics & 0.0 & 1.0 &
			\textbf{8.1558} & 10.6348 & 0.6560 &
			10.0467 & 12.9498 & 0.9938 &
			\textbf{9.9795} & 12.5423 & 1.0006 &
			9.9896 & 12.5433 & 1.0002 \\
			
			LSTM+Physics & 0.3 & 0.7 &
			8.1943 & 10.6364 & 0.6562 &
			\textbf{10.0305} & \textbf{12.9414} & \textbf{0.9925} &
			9.9807 & 12.5406 & 1.0004 &
			\textbf{9.9894} & \textbf{12.5433} & \textbf{1.0002} \\
			
			LSTM+Physics & 0.5 & 0.5 &
			8.1704 & 10.6322 & 0.6556 &
			10.0403 & 12.9432 & 0.9928 &
			9.9815 & \textbf{12.5398} & \textbf{1.0002} &
			9.9902 & 12.5435 & 1.0002 \\
			
			LSTM+Physics & 0.7 & 0.3 &
			8.1639 & \textbf{10.6237} & \textbf{0.6546} &
			10.0622 & 12.9576 & 0.9950 &
			10.0042 & 12.5447 & 1.0010 &
			9.9909 & 12.5438 & 1.0003 \\
			
			LSTM & 1.0 & 0.0 &
			8.1761 & 10.6289 & 0.6552 &
			10.0669 & 12.9638 & 0.9960 &
			9.9955 & 12.5401 & 1.0003 &
			9.9895 & 12.5433 & 1.0002 \\
			
			\bottomrule
		\end{tabular}
	}
\end{table*}

\begin{table*}[htbp]
	\centering
	\caption{
		Physics-guided LSTM performance for O$_{3}$ across forecasting horizons
		with different $\lambda_{\text{data}}$ and $\lambda_{\text{phys}}$ settings.
	}
	\label{tab:O3_lstm_physics_results}
	\vspace{2mm}
	
	\resizebox{\textwidth}{!}{
		\begin{tabular}{lccccccccccccccc}
			\toprule
			\textbf{Model} &
			$\lambda_{\text{data}}$ &
			$\lambda_{\text{phys}}$ &
			\multicolumn{3}{c}{\textbf{LAG 1}} &
			\multicolumn{3}{c}{\textbf{LAG 7}} &
			\multicolumn{3}{c}{\textbf{LAG 14}} &
			\multicolumn{3}{c}{\textbf{LAG 30}} \\
			
			\cmidrule(lr){4-6}
			\cmidrule(lr){7-9}
			\cmidrule(lr){10-12}
			\cmidrule(lr){13-15}
			
			& & &
			MAE & RMSE & NMSE &
			MAE & RMSE & NMSE &
			MAE & RMSE & NMSE &
			MAE & RMSE & NMSE \\
			
			\midrule
			LSTM+Physics & 0.0 & 1.0 &
			12.7986 & 19.4895 & 0.5954 &
			16.6970 & 25.2380 & 0.9939 &
			16.4360 & 23.9733 & 0.8927 &
			17.7256 & 25.3531 & 0.9856 \\
			
			LSTM+Physics & 0.3 & 0.7 &
			12.7856 & 19.5148 & 0.5970 &
			16.6655 & 25.2851 & 0.9976 &
			16.4991 & 23.9886 & 0.8938 &
			17.8042 & 25.3755 & 0.9874 \\
			
			LSTM+Physics & 0.5 & 0.5 &
			12.8550 & \textbf{19.4646} & \textbf{0.5939} &
			16.6616 & 25.2858 & 0.9977 &
			16.4206 & 23.9859 & 0.8936 &
			17.7970 & 25.3737 & 0.9872 \\
			
			LSTM+Physics & 0.7 & 0.3 &
			12.8602 & 19.4922 & 0.5956 &
			\textbf{16.6410} & 25.3056 & 0.9992 &
			16.4376 & 23.9728 & 0.8926 &
			17.7830 & 25.3708 & 0.9870 \\
			
			LSTM & 1.0 & 0.0 &
			\textbf{12.7642} & 19.4974 & 0.5959 &
			16.7570 & \textbf{25.2099} & \textbf{0.9917} &
			\textbf{16.3613} & \textbf{23.9691} & \textbf{0.8924} &
			\textbf{17.7549} & \textbf{25.3641} & \textbf{0.9865} \\
			
			\bottomrule
	\end{tabular}}
\end{table*}

\subsection{Summary of Key Findings}
For PM$_{2.5}$ at lag = 1, the LSTM+Physics (0.7, 0.3) achieves the best performance, at lag = 7, the LSTM+Physics (0.3, 0.7) achieves the lowest RMSE (12.9414) and NMSE (0.9925), and at lag = 14, the LSTM+Physics (0.5, 0.5) outperforms the rest. For lag = 30, the LSTM+Physics (0.3, 0.7) has the best overall performance. For the another pollutant O$_3$ at lag = 1, the LSTM+Physics (0.5, 0.5) achieves the best overall performance; at lag = 7, the baseline LSTM (1, 0) provides the lowest RMSE (25.2099) with competitive NMSE (0.9917); at lag = 14, the MLP+Physics (0, 1) yields the best performance and lag = 30, the MLP+Physics (0, 1) achieves the best overall results. From the observed results it could be stated that physics-guided deep-learning models improve the AQI forecasting accuracy. The LSTM+Physics model consistently performs best for PM$_{2.5}$ across all lags and also for short-term O$_3$ AQI prediction, while the MLP+Physics model generalizes better for long-term O$_3$ AQI forecasting. 

For the set of both pollutants and all horizons, there are four conclusions that can be drawn. First, the difficulty of the forecast rises with the horizon, and in general, the performance decreases from $LAG=1$ to $LAG=30$, which is exactly what we want to see in the target behavior in Figure~\ref{fig:pm_ozone_multihorizon}. Second, deep learning models (MLP and especially LSTM) generally are better than simple baselines. Third, physics guidance is the most unambiguous in terms of PM$_{2.5}$ particularly at short horizons, by making the stability of PM$_{2.5}$ more robust and enforcing physically meaningful pollutant-AQI consistency. Finally, LSTM-based models indicate the strongest performance over horizons, whereas the physical constraints can potentially allow further improvements in stability and interpretability based on the pollutant and horizon.

\section{Discussion}

This research study is an attempt to set an overall comprehensive benchmark of the classical, machine learning, deep learning, and knowledge-guided physics models for AQI multi-horizon forecasting in Texas's Dallas County. Overall, the experimental results highlight few significant insights regarding the experimented model’s behavior, the pollutant characteristics, and the role of knowledge guidance of physics.

First of all, the observed degradation in the overall performance for the AQI forecasting horizons with different lag increases is expected and in line with atmospheric dynamics. Short horizon AQI predictions have excellent temporal persistence, whereas longer time horizons have accumulated uncertainties due to meteorological variability and emission changes as well as nonlinear chemical interactions. This effect can be seen especially well in O$_3$, which has more day-to-day volatility than PM$_{2.5}$. Secondly, deep learning models perform invariably better than the classical model used to baseline for pollutants and AQI forecast horizons. It is worth mentioning that the MLP captures the nonlinear relationship between pollutant concentration and the AQI, and the LSTM captures nonlinear relationships further and models the temporal dependencies. The higher stability with longer horizons of LSTM models shows that sequence-based learning is particularly effective when it comes to AQI forecasting tasks. Third, an EPA breakpoint-based AQI physics model incorporates the learning process, enhancing the model's behavior beyond just pure accuracy metrics. For PM$_{2.5}$, physics guidance offered greatly reduced errors and more stable predictions, especially for short-time forecasts. The data indicates that PM$_{2.5}$ is related to AQI in a much more direct and stable manner, which could be well reinforced using physics-based constraints. On the other hand, straight AQI physics limitations are less useful for O$_3$ forecasting. A simple breakpoint-based AQI calculation is not enough to consider the complicated photochemical reactions, weather, and precursor emissions required for ozone production. However, through not relying on irrational forecasts and maintaining physically consistent AQI trajectories with time, physics-guided models still hold qualitative advantages.

Overall, it can be seen that if there is a strong and clear physical relationship between inputs and targets such as in the case of PM$_{2.5}$, physics-guided learning works best. Physics guiding plays less of a direct and improving accuracy role and more of a stabilizing and regularizing mechanism when the background process is more complicated, like it is in the case of O$_3$.

\section{Conclusion}

To enhance the accuracy and set the benchmark for the multi-horizon AQI forecasting in North Texas, Dallas County is represented in this research study by using EPA PM$_{2.5}$ and O$_3$ data. The experimental results proved that deep-learning models, particularly LSTM, consistently outperform simpler baselines across forecasting horizons, and incorporating EPA breakpoint-based AQI physics further improves prediction stability and physical plausibility, with the most noticeable benefits observed for PM$_{2.5}$ and short-term forecasts. For O$_3$, physics guidance contributes primarily to stabilizing predictions rather than large numerical error reductions. Overall, the findings demonstrate that lightweight physics guidance can enhance deep-learning–based AQI forecasting and that its effectiveness depends on pollutant characteristics. This benchmark provides practical guidance for selecting and deploying AQI forecasting models in urban air-quality applications, and considering meteorological parameters could enhance the performance of the physics-guided model.

\section*{Acknowledgments}
This research has been conducted in the Data Driven Decisions (D3) Lab in the Anuradha and Vikas Sinha Department of Data Science at the University of North Texas. 

\bibliographystyle{unsrt}  
\bibliography{references}  

\end{document}